\documentclass[11pt,a4paper]{article}
\usepackage[hyperref]{acl2021}
\usepackage{times}
\usepackage{latexsym}
\usepackage{bm}
\usepackage{booktabs}
\usepackage{graphicx}
\usepackage{subcaption}
\usepackage{tikz}
\newcommand*\circled[1]{\tikz[baseline=(char.base)]{
            \node[shape=circle,draw,inner sep=0.9pt] (char) {#1};}}
\usepackage{CJKutf8}
\usepackage{xspace}
\usepackage{amsfonts}
\usepackage{amsmath}
\usepackage{paralist}

\aclfinalcopy 
\def\aclpaperid{2134} 

\newcommand{\baselinea}{m-Transformer\xspace}
\newcommand{\method}{mRASP2\xspace}
\newcommand{\methodx}{mRASP2 w/o MC24 \xspace}
\newcommand{\methodl}{mRASP \xspace}
\newcommand{\methodb}{mRASP2 w/o AA \xspace}

\newcommand{\dataset}{PC32\xspace}

\newcommand{\mdataset}{MC24\xspace}
\newcommand{\tedm}{Ted-M\xspace}
\newcommand{\xx}{\mathbf{x}}
\newcommand{\yy}{\mathbf{y}}
\newcommand{\mml}{L}
\newcommand{\ml}{\mathcal{L}}
\newcommand{\mmd}{\mathcal{D}}

\title{Contrastive Learning for Many-to-many Multilingual Neural Machine Translation}

\author{ Xiao Pan, 
Mingxuan Wang, 
Liwei Wu, 
Lei Li \\
  ByteDance AI Lab \\
   \texttt{\{panxiao.94,wangmingxuan.89,wuliwei.000,lileilab\}@bytedance.com}
  }

\date{}

\begin{document}
\maketitle
\begin{abstract}

Existing multilingual machine translation approaches mainly focus on English-centric directions, while the non-English directions still lag behind. 
In this work, we aim to build a many-to-many translation system with an emphasis on the quality of non-English language directions.
Our intuition is based on the hypothesis that a universal cross-language representation  leads to better multilingual translation performance. 
To this end, we propose \method, a training method to obtain a single unified multilingual translation model. 
\method is empowered by two techniques:
\begin{inparaenum}[\it a)]
\item a contrastive learning scheme to close the gap among representations of different languages, and
\item data augmentation on both multiple parallel and monolingual data to further align token representations. 
\end{inparaenum}
For English-centric directions, \method  outperforms existing best unified model and achieves competitive or even better performance than the pre-trained and fine-tuned model mBART on tens of WMT's translation directions. 
For non-English directions, \method achieves an improvement of average 10+ BLEU compared with the multilingual Transformer baseline. 
Code, data and trained models are available at \url{https://github.com/PANXiao1994/mRASP2}. 
\end{abstract}


\section{Introduction}
\label{sec:intro}

Transformer~\cite{vaswani2017} has achieved decent performance for machine translation with rich bilingual parallel corpora. 
Recent work on multilingual machine translation aims to create a single unified model to translate many languages~\cite{johnson-etal-2017-googles,aharoni2019massively,zhang-2020-improving-massive,fan2020beyond,siddhant2020leveraging}.  
Multilingual translation models are appealing for two reasons. 
First, they are model efficient, enabling easier deployment~\cite{johnson-etal-2017-googles}. Further,  parameter sharing across different languages 
encourages knowledge transfer, which benefits low-resource translation directions and potentially enables zero-shot translation (i.e. direct translation between a language pair not seen during training)~\cite{ha2017effective,gu2019improved,ji2020cross}.

\begin{figure}[!t]
\centering
\includegraphics[width=1.0\columnwidth]{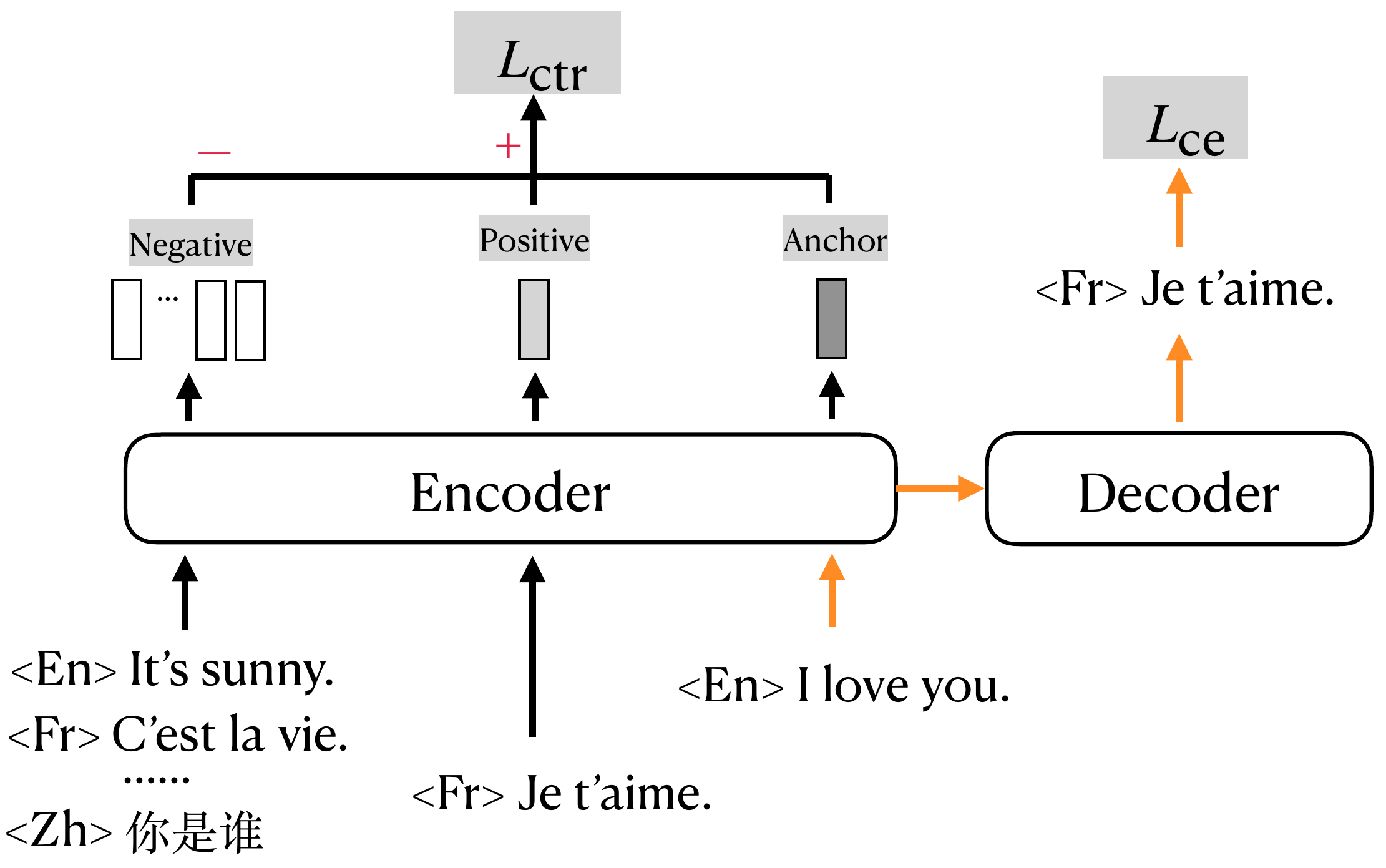}
\caption{The proposed \method. It takes a  pair of parallel sentences (or augmented pseudo-pair) and computes normal cross entropy loss with a multi-lingual encoder-decoder.
In addition, it computes contrastive loss on the representations of the aligned pair (positive example) and randomly selected non-aligned pair (negative example). } 
\label{fig:diagram}
\end{figure}

Despite these benefits, challenges still remain in multilingual NMT. 
First,  previous work on multilingual NMT does not always perform well as their corresponding bilingual baseline especially on rich resource language pairs~\cite{tan2019multilingual,zhang-2020-improving-massive,fan2020beyond}. 
Such performance gap becomes larger with the increasing number of  accommodated  languages for multilingual NMT, as model capacity necessarily must be split between many languages~\cite{arivazhagan2019massively}. 
In addition, an optimal setting for multilingual NMT should be effective for any language pairs, while most previous work focus on improving English-centric\footnote{``English-centric'' means that having English as the source or target language} directions~\cite{johnson-etal-2017-googles,aharoni2019massively,zhang-2020-improving-massive}.  
A few recent exceptions are ~\citet{zhang-2020-improving-massive} and ~\citet{fan2020beyond}, who trained many-to-many systems with introducing more non-English corpora, through data mining or back translation. 

In this work,  we take a step towards a unified many-to-many multilingual NMT with only English-centric parallel corpora and additional monolingual corpora. 
Our key insight is to close the representation gap between different languages to encourage transfer learning as much as possible.

As such, many-to-many translations can make the most of the knowledge from all supervised directions and the model can perform well for both English-centric and non-English settings. 
In this paper, we propose a multilingual COntrastive Learning framework for Translation (mCOLT or \method) to reduce the representation gap of different languages, as shown in Figure~\ref{fig:diagram}.

The objective of \method  ensures the model to represent similar sentences across languages in a shared space by training the encoder to minimize the representation distance of similar sentences.
In addition, we also boost \method by leveraging monolingual data to further improve multilingual translation quality. 
We introduce an effective aligned augmentation technique by extending RAS~\cite{mrasp} -- on both parallel and monolingual corpora to create pseudo-pairs. 
These pseudo-pairs are combined with multilingual parallel corpora in a unified training framework.

Simple yet effective, \method achieves  consistent translation performance improvements for both English-centric and non-English directions on a wide range of benchmarks.  
For English-centric directions, \method outperforms a strong multilingual baseline in 20 translation directions on WMT testsets. 
On 10 WMT translation benchmarks, \method even obtains  better results than the strong bilingual  mBART model.  
For zero-shot and unsupervised  directions, \method obtains surprisingly strong results on 36 translation directions\footnote{6 unsupervised directions $+$ 30 zero-shot directions}, with 10+ BLEU improvements on average.

\section{Methodology}
\label{sec:approach}
\method unifies both parallel corpora and monolingual corpora with contrastive learning. 
This section will explain our proposed \method. The overall framework is illustrated in Figure~\ref{fig:diagram}

\begin{figure*}[ht]
\centering
\begin{subfigure}[b]{0.48\linewidth}
         \centering
         \includegraphics[width=\textwidth]{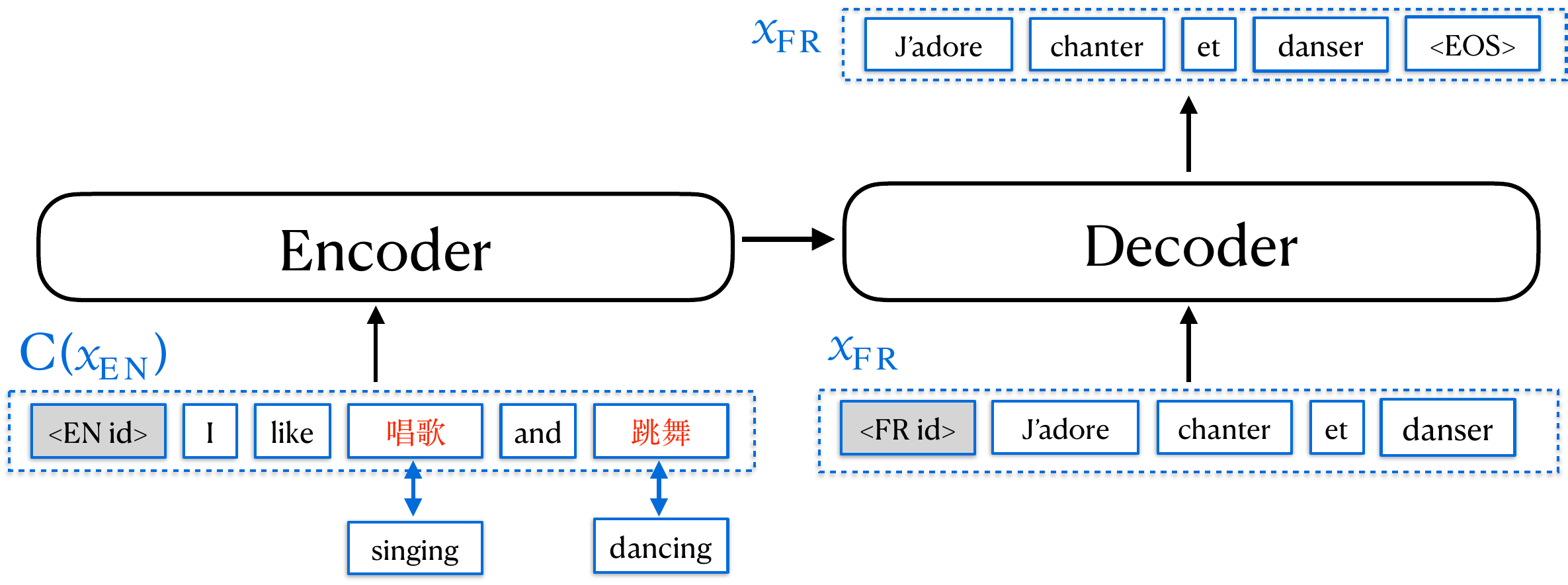}
         \subcaption{AA for Parallel Corpora}
         \label{fig:ras1}
     \end{subfigure}
     \hfill
     \begin{subfigure}[b]{0.48\linewidth}
         \centering
         \includegraphics[width=\textwidth]{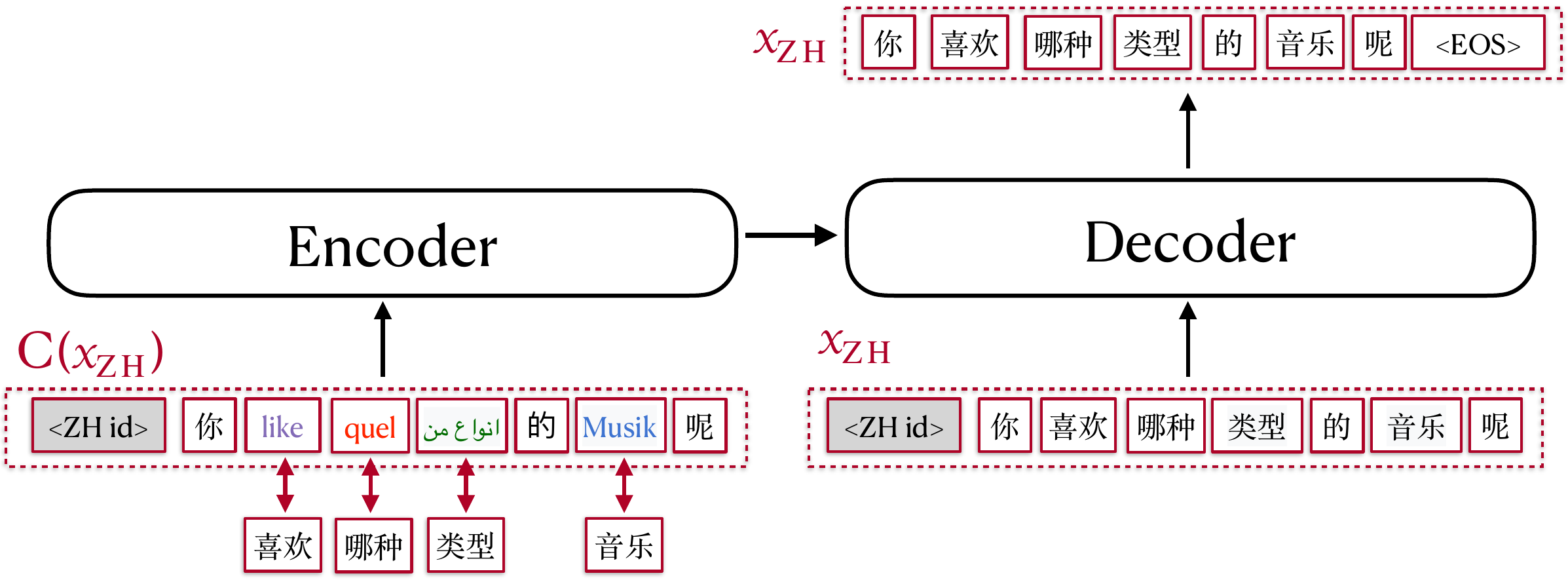}
         \caption{AA for Monolingual Corpora}
         \label{fig:ras2}
     \end{subfigure}
\caption{
Aligned augmentation on both parallel and monolingual data by replacing words with the same meaning in synonym dictionaries. It either creates a pseudo-parallel example (left) or a pseudo self-parallel example (right).}
\label{fig:ras}
\end{figure*}

\subsection{Multilingual Transformer}
A multilingual neural machine translation model learns a many-to-many mapping function $f$ to translate from one language to another.   
To distinguish different languages, we add an additional language identification token preceding each sentence, for both source side and target side.
The base architecture of \method is the state-of-the-art Transformer \cite{vaswani2017}.
A little different from previous work, we choose a  larger setting with a 12-layer encoder and a 12-layer decoder to increase the model capacity. The model dimension is 1024 on 16 heads. 
To ease the training of the deep model, we apply Layer Normalization for word embedding and pre-norm residual connection following \citet{deep-transformer} for both encoder and decoder. 
Therefore, our multilingual NMT baseline is much stronger than that of Transformer big model. 

More formally, we define  $\mml=\left\{\mml_{1}, \ldots, \mml_{M}\right\}$ where $\mml$ is a collection of M languages involving in the training phase. 
$\mmd_{i,j}$ denotes a parallel dataset of $(\mml_{i}, \mml_j)$, and $\mmd$ denotes all parallel datasets. The training loss is cross entropy  defined as:
\begin{equation}
    \ml_\text{{ce}} = \sum_{\xx^i,\xx^j\in \mmd}  -\log P_\theta(\xx^i|\xx^j)
\label{eq:multi}
\end{equation}
where $\xx^i$ represents a sentence in language $\mml_i$, and $\theta$ is the parameter of multilingual Transformer model. 

\subsection{Multilingual Contrastive Learning}
Multilingual Transformer enables implicitly learning shared representation of different languages.  \method introduces contrastive loss to explicitly bring different languages to map a shared semantic space. 




The key idea of contrastive learning is to minimize the representation gap of similar sentences and maximize that of irrelevant sentences. Formally,
given a bilingual translation pairs $(\xx^i, \xx^j) \in \mmd$, $(\xx^i, \xx^j)$ is the positive example and we randomly choose a sentence $\yy^j$ from language $L_j$ to form a negative example\footnote{It is possible that $L_j$ = $L_i$ } $(\xx^i, \yy^j)$. The objective of contrastive learning is to minimize the following loss:
\begin{equation}
    \ml_\text{ctr} = -\sum_{\xx_i,\xx_j\in \mmd} \log \frac{e^{\text{sim}^+(\mathcal{R}(\xx^i), \mathcal{R}(\xx^j))/\tau}}{\sum_{\yy^j} e^{\text{sim}^-(\mathcal{R}(\xx^i), \mathcal{R}(\yy^j))/\tau }}
    \label{eq:cl}
\end{equation} 
where $\text{sim}(\cdot)$ calculates the similarity of different sentences. $+$ and $-$ denotes positive and negative respectively. $\mathcal{R}(s)$ denotes the average-pooled encoded output of an arbitrary sentence $s$. $\tau$ is the temperature, which controls the difficulty of distinguishing between positive and negative examples\footnote{Higher temperature increases the difficulty to distinguish positive sample from negative ones.}. In our experiments, it is set to $0.1$. 
The similarity of two sentences is calculated with the cosine similarity
of  the average-pooled encoded output. 
To simplify implementation,  the negative samples are sampled from the same training batch.
Intuitively,  by  maximizing  the  softmax  term $\text{sim}^+(\mathcal{R}(\xx^i), \mathcal{R}(\xx^j))$, the contrastive loss forces their  semantic  representations  projected  close to each other. In the meantime,   the softmax function  also  minimizes  the  non-matched  pairs $\text{sim}^-(\mathcal{R}(\xx^i), \mathcal{R}(\yy^j))$.



During the training of \method, the model can be optimized by jointly minimizing the contrastive training loss and translation loss:
\begin{equation}
\label{eq:all}
    \ml = \ml_{\text{ce}} + \lambda |s| \ml_{\text{ctr}}
\end{equation}
where $\lambda$ is the coefficient to balance the two training losses. Since $\ml_{\text{ctr}}$ is calculated on the sentence-level and $\ml_{\text{ce}}$ is calculated on the token-level, therefore $\ml_{\text{ctr}}$ should be multiplied by the averaged sequence length $|s|$.

\subsection{Aligned Augmentation}

We then will introduce how to improve \method with data augmentation methods, including the introduction of noised bilingual and noised monolingual data for multilingual NMT. 
The above two types of training samples are illustrated in Figure~\ref{fig:ras}.


\citet{mrasp} propose Random Aligned Substitution technique (or RAS\footnote{They apply RAS only on parallel data}) that builds code-switched sentence pairs ($C(\mathbf{x}^i), \mathbf{x}^j$) for multilingual pre-training. 
In this paper, we extend it to Aligned Augmentation (AA), which can also be applied to monolingual data. 

For a bilingual or monolingual sentence pair ($\xx^i$, $\xx^j$)\footnote{$\xx^i$ is in language $\mml_i$ and $\xx^j$ is in language $\mml_j$, where $i,j \in \left\{\mml_{1}, \ldots, \mml_{M}\right\}$}, AA creates a perturbed sentence $C(\xx^i)$ by replacing aligned words from a synonym dictionary\footnote{We will release our synonym dictionary}. For every word contained in the synonym dictionary, we randomly replace it to one of its synonym with a probability of 90\%.

 For a bilingual sentence pair $(\xx^i,\xx^j)$, AA creates a pseudo-parallel training example $(C(\xx^i),\xx^j)$. For monolingual data, AA takes a sentence $\xx^i$ and generates its perturbed $C(\xx^i)$ to form a pseudo self-parallel example $(C(\xx^i), \xx^i)$. $(C(\xx^i),\xx^j)$ and $(C(\xx^i), \xx^i)$ is then used in the training by calculating both the translation loss and contrastive loss. For a pseudo self-parallel example $(C(\xx^i), \xx^i)$, the translation loss is basically the reconstruction loss from the perturbed sentence to the original one.

\section{Experiments}
\label{sec:exps}

\begin{table*}[tb]
\begin{center}
\resizebox{\linewidth}{!}{
\begin{tabular}{lllllllllll|rr}
\toprule
&
  \multicolumn{2}{c}{\begin{tabular}[c]{@{}c@{}}En-Fr\\ wmt14\end{tabular}} &
  \multicolumn{2}{c}{\begin{tabular}[c]{@{}c@{}}En-Tr\\ wmt17\end{tabular}} &
  \multicolumn{2}{c}{\begin{tabular}[c]{@{}c@{}}En-Es\\ wmt13\end{tabular}} &
  \multicolumn{2}{c}{\begin{tabular}[c]{@{}c@{}}En-Ro \\ wmt16\end{tabular}} &
  \multicolumn{2}{c}{\begin{tabular}[c]{@{}c@{}}En-Fi\\ wmt17\end{tabular}} &
  \multicolumn{1}{c}{Avg} &
  \multicolumn{1}{c}{$\Delta$}
   \\
 &
  \multicolumn{1}{c}{$\rightarrow$} &
  \multicolumn{1}{c}{$\leftarrow$} &
  \multicolumn{1}{c}{$\rightarrow$} &
  \multicolumn{1}{c}{$\leftarrow$} &
  \multicolumn{1}{c}{$\rightarrow$} &
  \multicolumn{1}{c}{$\leftarrow$} &
  \multicolumn{1}{c}{$\rightarrow$(*)} &
  \multicolumn{1}{c}{$\leftarrow$} &
  \multicolumn{1}{c}{$\rightarrow$} &
  \multicolumn{1}{c}{$\leftarrow$} &
   &
   \\ \midrule
   \it{bilingual} &&&&&&&&&&&& \\
   Transformer-6\cite{mrasp} &
  43.2 &
  39.8 &
  - &
  - &
  - &
  - &
  34.3 &
  34.0 &
  - &
  - &
  \multicolumn{1}{c}{-} & \\
   Transformer-12\cite{liu2020multilingual} &
  41.4 &
  - &
  9.5 &
  12.2 &
  33.2 &
  - &
  34.3 &
  36.8 &
  20.2 &
  21.8 &
  \multicolumn{1}{c}{-} & \\
  \hline
  \it{pre-train \& fine-tuned} &&&&&&&&&&&& \\

    Adapter~\cite{bapna2019simple} &
  - &
  - &
  - &
  - &
  \textbf{35.4} &
  33.7 &
  - &
  - &
  - &
  - &
  \multicolumn{1}{c}{-} &
  \\ 
  
  mBART\cite{liu2020multilingual} &
  41.1 &
  - &
  17.8 &
  22.5 &
  34.0 &
  - &
  37.7 &
  38.8 &
  22.4 &
  28.5 &
  \multicolumn{1}{c}{-} & \\
  
  XLM\cite{xlm} &
  - &
  - &
  - &
  - &
  - &
  - &
  - &
  38.5 &
  - &
  - &
  \multicolumn{1}{c}{-} & \\
  
  MASS\cite{song2019mass} &
  - &
  - &
  - &
  - &
  - &
  - &
  - &
  39.1 &
  - &
  - &
  \multicolumn{1}{c}{-} & \\
  
  mRASP\cite{mrasp} &
  \textbf{44.3} &
  \textbf{45.4} &
  20.0 &
  23.4 &
  - &
  - &
  37.6 &
  38.9 &
  \textbf{24.0} &
  28.0 &
  \multicolumn{1}{c}{-} &
  \\ 
  
   \hline
  \it{unified multilingual} &&&&&&&&&&&& \\
 
   Multi-Distillation ~\cite{tan2019multilingual} &
  - &
  - &
  - &
  - &
  - &
  - &
  31.6 &
  35.8 &
  22.0 &
  21.2 &
  \multicolumn{1}{c}{-} & \\

\baselinea &
  42.0 & 38.1 & 18.8 & 23.1 & 32.8 & 33.7 & 35.9 & 37.7 & 20.0 & 28.2 &
  \multicolumn{1}{c}{31.03} &
   \\
\methodl w/o finetune(**) & 43.1 & 39.2 & 20.0 & 25.2 & 34.0 & 34.3 & 37.5 & 38.8 & 22.0 & 29.2 & 32.33 & +1.30  \\
\method &
  43.5 & 39.3 & \textbf{21.4} & \textbf{25.8} & 34.5 & \textbf{35.0} & \textbf{38.0} & \textbf{39.1} & 23.4 & \textbf{30.1} &
  \multicolumn{1}{c}{\textbf{33.01}} &
  \multicolumn{1}{c}{\textbf{+1.98}} 
  \\ \bottomrule 
 
\end{tabular}
}
\caption{ Performance (tokenized BLEU) on WMT \textbf{supervised} translation directions. Consistent BLEU gains are observed in 20 directions (See Appendix) and in this table we pick the representative ones. Different from our work, final BLEU scores of mBART, XLM, MASS and \methodl are obtained by  multilingual pre-training and \textbf{fine-tuning} on a single direction. Adapter is a trade-off between unified multilingual model and bilingual model (trained on 6 languages on WMT data).  Multi-Distillation is improved over Adapter with selective distillation methods. 
Results for Transformer-6 (6 layers for encoder and decoder) are from \citet{mrasp}. 
Results for Transformer-12 (12 layers for encoder and decoder separately) are from \citet{liu2020multilingual}. 
(*) Note that for En$\rightarrow$Ro direction, we follow the previous setting to calculate BLEU score after removing Romanian dialects. (**) For \methodl w/o finetune we report the results implemented by ourselves, with 12 layers encoder and decoder and our data.
Both \baselinea and 
our \method have 12 layers for encoder and decoder. 
}
\label{tab:supa}
\end{center}
\end{table*}

This section shows that \method can achieve substantial improvements over previous many-to-many multilingual translation on a wide range of benchmarks. 
Especially, it obtains substantial  gains on zero-shot directions.

\subsection{Settings and Datasets}

\paragraph{Parallel Dataset \dataset} 
We use the parallel dataset \dataset provided by \citet{mrasp}.
It contains a large public parallel corpora of 32 English-centric language pairs. 
The total number of sentence pairs is 97.6 million. 

We apply AA on \dataset by randomly replacing words in the source side sentences with synonyms from an arbitrary bilingual dictionary provided by \cite{lample2017unsupervised}\footnote{https://github.com/facebookresearch/MUSE}. For words in the dictionaries, we replace them into one of the synonyms with a probability of 90\% and keep them unchanged otherwise. We apply this augmentation in the pre-processing step before training.

\paragraph{Monolingual Dataset \mdataset} 
We create a dataset \mdataset with monolingual text in 24 languages\footnote{Bg, Cs, De, El, En, Es, Et, Fi, Fr, Gu, Hi, It, Ja, Kk, Lt, Lv, Ro, Ru, Sr, Tr, Zh, Nl, Pl, Pt}. 
It is a subset of the Newscrawl\footnote{http://data.statmt.org/news-crawl} dataset by retaining only those languages in \dataset, plus three additional languages that are not in \dataset (Nl, Pl, Pt). 
In order to balance the volume across different languages, we apply temperature sampling $\tilde{n_i} = \left(n_i / \sum_j n_j\right)^{1/T}$ with $T$=5 over the dataset, where $n_i$ is the number of sentences in $i$-th language. Then we apply AA on monolingual data. The total number of sentences in \mdataset is 1.01 billion. The detail of data volume is listed in the Appendix.

\begin{table*}[!tb]
\begin{center}
\begin{tabular}{lllllll|ll|rr}
\toprule
 &
  \multicolumn{2}{c}{\begin{tabular}[c]{@{}c@{}}En-Nl\\ iwslt2014\end{tabular}} &
  \multicolumn{2}{c}{\begin{tabular}[c]{@{}c@{}}En-Pt\\ opus-100\end{tabular}} &
  \multicolumn{2}{c}{\begin{tabular}[c]{@{}c@{}}En-Pl\\ wmt20\end{tabular}} &
  \multicolumn{2}{c}{\begin{tabular}[c]{@{}c@{}}Nl-Pt\\ -\end{tabular}} &
  \multicolumn{1}{c}{Avg} &
  \multicolumn{1}{c}{$\Delta$} \\ 
 &
  \multicolumn{1}{c}{$\rightarrow$} &
  \multicolumn{1}{c}{$\leftarrow$} &
  \multicolumn{1}{c}{$\rightarrow$} &
  \multicolumn{1}{c}{$\leftarrow$} &
  \multicolumn{1}{c}{$\rightarrow$} &
  \multicolumn{1}{c}{$\leftarrow$} &
  \multicolumn{1}{c}{$\rightarrow$} &
  \multicolumn{1}{c}{$\leftarrow$} &
   &
   \\ \hline
\baselinea & 1.3          & 7.0             & 3.7           & 10.7          & 0.6          & 3.2 & - & -         & 4.42  &      \\
\methodl & 0.7 & 10.6 & 3.7 & 11.6 & 0.5 & 5.3 & - & -   & 5.40  &  +0.98    \\
\method & \textbf{10.1} & \textbf{28.5} & \textbf{18.4} & \textbf{30.5} & \textbf{6.7} & \textbf{17.1} & \textbf{9.3} & \textbf{8.3} & \textbf{18.55} & \textbf{+14.13} \\
\bottomrule
\end{tabular}
\caption{ \method outperforms \baselinea in \textbf{unsupervised} translation directions by a large margin. We report tokenized BLEU above. For Nl$\leftrightarrow$Pt, \method achieves reasonable results after trained only on monolingual data of both sides. The averaged score is calculated without the Nl$\leftrightarrow$Pt directions. }
\label{tab:supc}
\end{center}
\end{table*}

\begin{table*}[!tb]
\begin{center}
\resizebox{0.75\linewidth}{!}{
\begin{tabular}{lllllllr}
\toprule
 &
  \multicolumn{2}{c}{Ar} &
  \multicolumn{2}{c}{Zh} &
  \multicolumn{2}{c}{Nl(*)} &
  \multicolumn{1}{c}{} \\
 &
  \multicolumn{1}{c}{X$\rightarrow$Ar} &
  \multicolumn{1}{c}{Ar$\rightarrow$X} &
  \multicolumn{1}{c}{X$\rightarrow$Zh} &
  \multicolumn{1}{c}{Zh$\rightarrow$X} &
  \multicolumn{1}{c}{X$\rightarrow$Nl} &
  \multicolumn{1}{c}{Nl$\rightarrow$X} &
  \multicolumn{1}{c}{} \\
  \hline
  Pivot    & 5.5           & 17.0          & 28.5          & 16.4     & 2.2   & 6.0    &  \multicolumn{1}{c}{} \\ \hline
\baselinea & 3.7           & 5.6  & 6.7  & 4.1  & 2.3          & \textbf{6.3}         & \multicolumn{1}{c}{}  \\
\method   & \textbf{5.3}     & \textbf{17.3}    & \textbf{29.0}    & \textbf{14.5}    & \textbf{5.3} & 6.1          &   \multicolumn{1}{c}{} 
\\ \midrule
 &
  \multicolumn{2}{c}{Fr} &
  \multicolumn{2}{c}{De} &
  \multicolumn{2}{c}{Ru} &
  Avg of all  \\
 &
  \multicolumn{1}{c}{X$\rightarrow$Fr} &
  \multicolumn{1}{c}{Fr$\rightarrow$X} &
  \multicolumn{1}{c}{X$\rightarrow$De} &
  \multicolumn{1}{c}{De$\rightarrow$X} &
  \multicolumn{1}{c}{X$\rightarrow$Ru} &
  \multicolumn{1}{c}{Ru$\rightarrow$X} &
  \multicolumn{1}{c}{} \\
  \hline
  \hline
Pivot    & 26.1         & 22.3         & 14.4         & 14.2      &   16.6  &    19.9          &    15.56 \\ \hline
\baselinea     & 7.7           & 4.8           & 4.2           & 4.8           & 5.7          & 4.8          & 5.05  \\
\method &
  \textbf{23.6}    & \textbf{21.7}    & \textbf{12.3}    & \textbf{15.0}    & \textbf{16.4}         & \textbf{19.1} &
  \textbf{15.31} \\

  \bottomrule
\end{tabular}
}
\caption{\textbf{Zero-Shot:} We report de-tokenized BLEU using sacreBLEU in OPUS-100. We observe consistent BLEU gains in zero-shot directions on different evaluation sets, see Appendix for more details. \method further improves the quality. We also list BLEU of pivot-based model (X$\rightarrow$En then En$\rightarrow$Y using \baselinea) as a reference, \method only lags behind Pivot by -0.25 BLEU. (*) Note that Dutch(Nl) is not included in \dataset. }
\label{tab:zs}
\end{center}
\end{table*}

We apply AA on \mdataset by randomly replacing words in the source side sentences with synonyms from a multilingual dictionary. Therefore the source side might contain multiple language tokens (preserving the semantics of the original sentence), and the target is just the original sentence. The replace probability is also set to 90\%. We apply this augmentation in the pre-processing step before training. We will release the multilingual dictionary and the script for producing the noised monolingual dataset.

\paragraph{Evaluation Datasets}
For supervised directions, most of our evaluation datasets are from  WMT and IWSLT benchmarks, for pairs that are not available in WMT or IWSLT, we use OPUS-100 instead. 

For zero-shot directions, we follow ~\cite{zhang-2020-improving-massive} and use their proposed OPUS-100 zero-shot testset. The testset is comprised of 6 languages (Ru, De, Fr, Nl, Ar, Zh), resulting in 15 language pairs and 30 translation directions.

We report de-tokenized BLEU with SacreBLEU \cite{DBLP:conf/wmt/Post18}. For tokenized BLEU, we tokenize both reference and hypothesis using Sacremoses\footnote{{https://github.com/alvations/sacremoses}} toolkit then report BLEU using the \texttt{multi-bleu.pl} script\footnote{{https://github.com/moses-smt/mosesdecoder}}. 
For Chinese (Zh), BLEU score is calculated on character-level.

\paragraph{Experiment Details} 
We use the Transformer model in our experiments, with 12 encoder layers and 12 decoder layers. The embedding size and FFN dimension are set to 1024. 
We use dropout = 0.1, as well as a learning rate of 3e-4 with polynomial decay scheduling and a warm-up step of 10000. For optimization, we use Adam optimizer \cite{adam} with $\epsilon$ = 1e-6 and $\beta_2$ = 0.98. To stabilize training, we set the threshold of gradient norm to be 5.0 and clip all gradients with a larger norm.
We set the hyper-parameter  $\lambda=1.0$ in Eq.\ref{eq:all} during training.
For multilingual vocabulary, 
we follow the shared BPE \cite{bpe} vocabulary of \citet{mrasp}, which includes 59 languages. The vocabulary contains 64808 tokens. After adding 59  language tokens, the total size of vocabulary is 64867.



\section{Experiment Results}
This section shows that \method provides consistent performance gains for supervised and unsupervised English-centric translation directions as well as for non-English directions.

\subsection{English-Centric Directions}
\paragraph{Supervised Directions}

\begin{table*}[!ht]
\begin{center}
\resizebox{0.95\linewidth}{!}{
\begin{tabular}{l|l|lll|rrr}
\toprule
     & model  & CTL & AA  & \mdataset & Supervised & Unsupervised & Zero-shot \\
\hline

\circled{1}& \baselinea & & & & 28.65      & 4.42         & 5.05      \\
\circled{2}& \methodl w/o f.t.(*)   & &\checkmark & & 29.82 & 5.40 & 4.91 \\
\circled{3}& \methodb   & \checkmark & & & 28.79      & 4.75         & 13.55     \\
\circled{4}& \methodx   & \checkmark & \checkmark & & 29.96 & 5.80 & 14.60 \\
\circled{5}& \method    & \checkmark& \checkmark & \checkmark & \textbf{30.36}      &  \textbf{18.55}        &  \textbf{15.31}   \\
\bottomrule
\end{tabular}
}
\caption{Summary of average BLEU of \methodb and \method in different scenarios. We report averaged tokenized BLEU. For supervised translation, we report the average of 20 directions; for zero-shot translation, we report the average of 30 directions of OPUS-100.  \methodl excludes \mdataset and contrastive loss from \method. \methodb only adopts contrastive learning on the basis of \baselinea. \methodx excludes \mdataset from \method. (*) Note that results of \methodl are computed without fine-tuning. }\label{tab:summ}
\end{center}
\end{table*}

As shown in Table ~\ref{tab:supa}, \method clearly improves multilingual baselines by a large margin in 10 translation directions. 
Previously, multilingual machine translation underperforms bilingual translation in rich-resource scenarios. It is worth noting that our multilingual machine translation baseline is very competitive. 
It is even on par with the strong mBART bilingual model, which is fine-tuned on a large scale unlabeled monolingual dataset. 
\method further improves the performance. 
  
We summarize the key factors for the success  training of  our baseline\footnote{many-to-many Transformer trained on \dataset as in \citet{johnson-etal-2017-googles} except that we apply language indicator the same way as \citet{fan2020beyond}}  \baselinea:
\begin{inparaenum}[a)]
\setlength{\itemsep}{2pt}
\setlength{\parsep}{2pt}
\setlength{\parskip}{2pt}
    \item The batch size plays  a crucial role in the success of training multilingual NMT. We use $8\times4$ NVIDIA V100 with update frequency 50 to train the models and each batch contains about 3 million tokens. 
    \item We enlarge the number of layers from 6 to 12  and observe  significant improvements for multilingual NMT. By contrast, the gains from increasing the bilingual model size is not that large. mBART also uses 12 encoder and decoder layers. 
    \item We use gradient norm to stable the  training. Without this regularization, the large scale training will collapse sometimes.    
\end{inparaenum}

\paragraph{Unsupervised Directions}
In Table~\ref{tab:supc}, we observe that \method achieves reasonable results on unsupervised translation directions. The language pairs of En-Nl, En-Pt, and En-Pl are never observed by \baselinea. 
\baselinea sometimes achieves reasonable BLEU for X$\rightarrow$En, e.g. $10.7$ for Pt$\rightarrow$En, since there are many similar languages in \dataset, such as Es and Fr. Not surprisingly, it totally fails on En$\rightarrow$X directions.
By contrast, \method obtains +14.13 BLEU score on an average without explicitly introducing  supervision signals for these directions. 

Furthermore, \method achieves reasonable BLEU scores on Nl$\leftrightarrow$Pt directions even though it has only been trained on monolingual data of both sides. This indicates that by simply incorporating monolingual data with parallel data in the unified framework, \method successfully enables unsupervised translation through its unified multilingual representation.

\subsection{Zero-shot Translation for non-English Directions}
Zero-shot Translation has been an intriguing topic in multilingual neural machine translation. Previous work shows that the multilingual NMT model can do zero-shot translation directly. However, the translation quality is quite poor compared with pivot-based model.

We evaluate \method on the OPUS-100 \cite{zhang-2020-improving-massive} zero-shot test set, which contains 6 languages\footnote{Arabic, Chinese, Dutch, French, German, Russian} and 30 translation directions in total. 
To make the comparison  clear, we also report the results of several different baselines.
\methodb only adopt contrastive learning on the basis of \baselinea. 
\methodx excludes monolingual data from \method.  

The evaluation results are listed in Appendix and we summarize them in Table~\ref{tab:zs}. 
We find that our \method significantly outperforms \baselinea and substantially narrows the gap with pivot-based model. This is in line with our intuition that bridging the representation gap of different languages can improve the zero-shot translation. 

The main reason is that contrastive loss, aligned augmentation and additional monolingual data enable a better language-agnostic sentence representation.  
It is worth noting that, \citet{zhang-2020-improving-massive} achieves BLEU score improvements on zero-shot translations at sacrifice of about 0.5 BLEU score loss on English-centric directions. By contrast, \method improves zero-shot translation by a large margin without losing performance on English-Centric directions. 
Therefore, \method has a great potential to serve many-to-many translations, including both English-centric and non-English directions.




\section{Analysis}
\label{sec: analysis}

To understand what contributes to the performance gain, we conduct analytical experiments in this section.  First we summarize and analyze the performance of \method in different scenarios. Second we adopt the sentence representation of \method to retrieve similar sentences across languages. This is to verify our argument that the improvements come from the universal language representation learned by \method. Finally we visualize the sentence representations, \method indeed draws the representations closer.


\subsection{Ablation Study}
To make a better understanding of the effectiveness of \method, we evaluate models of different settings. We summarize the experiment results in Table~\ref{tab:summ}:

\begin{itemize}
\item \circled{1}v.s.\circled{3}: \circled{3} performs comparably with \baselinea in supervised and unsupervised scenarios, whereas achieves a substantial BLEU improvement for zero-shot translation. This indicates that by introducing contrastive loss, we can improve zero-shot translation quality without harming other directions. 
\item \circled{2}v.s.\circled{4}: \circled{2} performs poorly for zero-shot directions. This means contrastive loss is crucial for the performance in zero-shot directions.
\item \circled{5}: \method further improves BLEU in all of the three scenarios, especially in unsupervised directions. Therefore it is safe to conjecture that by accomplishing with monolingual data, \method learns a better representation space.
\end{itemize}

\begin{table*}[t]
\begin{center}
\begin{tabular}{lllllllllll}
\toprule
Lang       & Fr   & De   & Zh   & Ro   & Cs                   & Tr             & Ru            & NL& PL   & Pt                   \\ \midrule
\baselinea & 91.7 & 96.8 & 87.0 & 90.6 & 84.8                & 91.1          & 89.1         & 25.6 & 6.3 & 37.3                \\
\methodb    & 91.7 & 97.3 & 89.9 & 91.4 & 86.1                & 92.4          & 90.4         & 35.7 & 14.3 & 46.5                \\
\method &
  \textbf{93.0} &
  \textbf{98.0} &
  \textbf{90.7} &
  \textbf{91.9} &
  \textbf{89.3} &
  \textbf{92.4} &
  \textbf{92.3} &
  \textbf{60.3} &
  \textbf{28.1} &
  \textbf{58.6} \\

\bottomrule 
\end{tabular}
\caption{\textbf{English-Centric:} Sentence retrieval top-1 accuracy on Tatoeba evaluation set. The reported accuracy is the average of En$\rightarrow$X and X$\rightarrow$En accuracy. \method outperforms \baselinea on all directions in English-centric sentence retrieval task.}
\label{tab:tatoeba}
\end{center}
\end{table*}

\subsection{Similarity Search}
In order to verify whether \method learns a better representation space, we conduct a set of similarity search experiments.
Similarity search is a task to find the nearest neighbor of each sentence in another language according to cosine similarity. We argue that \method benefits this task in the sense that it bridges the representation gap across languages. Therefore we use the accuracy of similarity search tasks as a quantitative indicator of cross-lingual representation alignment.

We conducted comprehensive experiments to support our argument and experiment on \method and \methodb.We divide the experiments into two scenarios: First we evaluate our method on Tatoeba dataset \cite{tatoeba}, which is English-centric. Then we conduct similar similarity search task on non-English language pairs. Following \citet{cross-iter}, we construct a multi-way parallel testset (\tedm) of 2284 samples by filtering the test split of ted\footnote{{http://phontron.com/data/ted\_talks.tar.gz}} that have translations for all 15 languages\footnote{Arabic, Czech, German, English, Spanish, French, Italian, Japanese, Korean, Dutch, Romanian, Russian, Turkish, Vietnamese, Chinese}.

Under both settings, we follow the same strategy: We use the average-pooled encoded output as the sentence representation. For each sentence from the source language, we search the closest sentence in the target set according to cosine similarity.

\paragraph{English-Centric: Tatoeba}
We display the evaluation results in Table~\ref{tab:tatoeba}. We detect two trends: (i) The overall accuracy follows the rule: \baselinea $ <$ \methodb $ <$ \method. (ii) \method brings more significant improvements for languages with less data volume in \dataset. The two trends mean that \method increases translation BLEU score in a sense that it bridges the representation gap across languages.

\begin{figure}[t]
     \centering
     \begin{subfigure}[b]{0.45\linewidth}
         \centering
         \includegraphics[width=\linewidth]{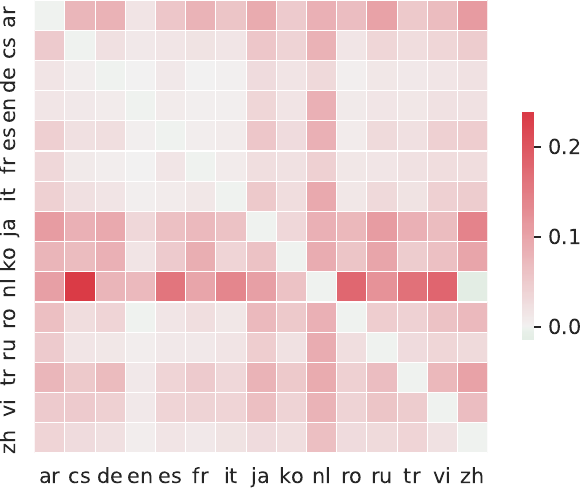}
         \subcaption{$\Delta$acc of \methodb over \baselinea}
         \label{fig:compa}
     \end{subfigure}
     \hfill
     \begin{subfigure}[b]{0.45\linewidth}
         \centering
         \includegraphics[width=\linewidth]{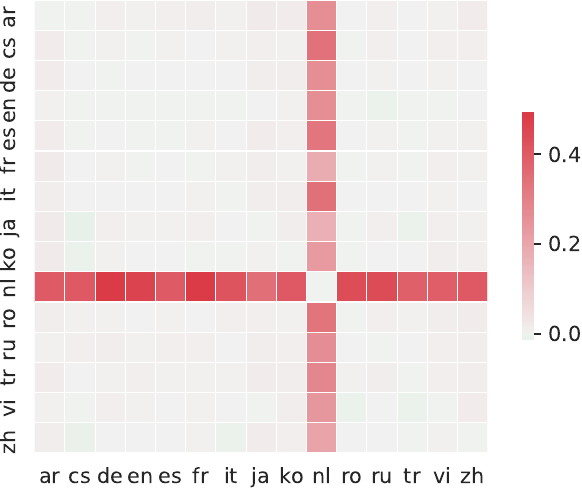}
         \subcaption{$\Delta$acc of \method over \methodb}
         \label{fig:compb}
     \end{subfigure}
        \caption{Accuracy Improvements of \baselinea \textbf{$\rightarrow$} \methodb \textbf{$\rightarrow$} \method for \tedm. Darker red means larger improvements. \methodb generally improves accuracy over \baselinea and \method especially improves the accuracy X $\leftrightarrow$ Nl over \methodb.}
        \label{fig:tedacc}
\end{figure}

\paragraph{Non-English: \tedm}

\begin{table}[!t]
\begin{center}
\begin{tabular}{l|l|l}
\toprule
           & Top1 Acc & $\Delta$  \\
   \hline
\baselinea & 79.8       & -      \\
\methodb    & 84.4       & +4.8      \\
\method   & \textbf{89.6}   & +9.8   \\
\bottomrule
\end{tabular}
\caption{\textbf{Non-English:} The averaged sentence similarity search top-1 accuracy on \tedm testset. \baselinea $<$ \methodb $<$ \method, which is consistent with the results in English-centric scenario.}

\label{tab:ted}
\end{center}
\end{table}

It will be more convincing to argue that \method indeed bridges the representation gap if similarity search accuracy increases on zero-shot directions.
We list the averaged top-1 accuracy of 210 non-English directions\footnote{15 languages, resulting in 210 directions} in Table~\ref{tab:ted}. The results show that \method increases the similarity search accuracy in zero-shot scenario. The results support our argument that our method generally narrows the representation gap across languages.

To better understanding the specifics beyond the averaged accuracy, we plot the accuracy improvements in the heat map in Figure~\ref{fig:tedacc}. \methodb brings general improvements over \baselinea. \method especially improves on Dutch(Nl). This is because \method introduces monolingual data of Dutch while \methodb includes no Dutch data.

\begin{figure*}[ht]
     \centering
     \begin{subfigure}[b]{0.42\linewidth}
         \centering
         \includegraphics[width=\textwidth]{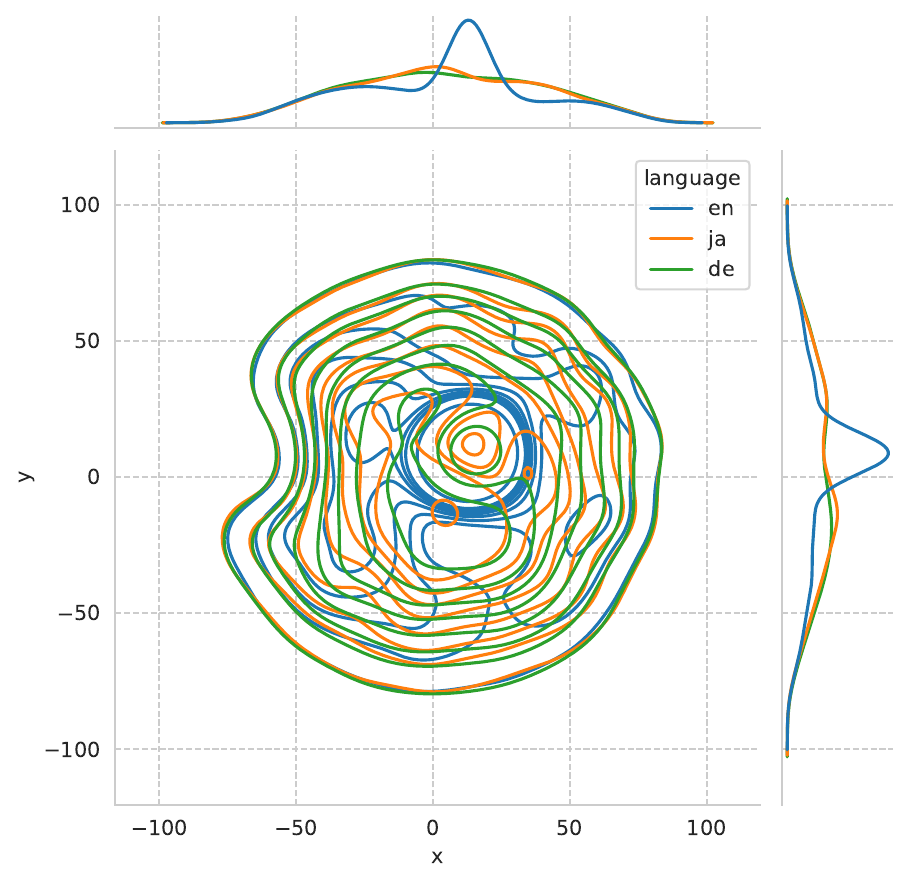}
         \subcaption{\baselinea}
         \label{fig:visa}
     \end{subfigure}
     \hfill
     \begin{subfigure}[b]{0.42\linewidth}
         \centering
         \includegraphics[width=\textwidth]{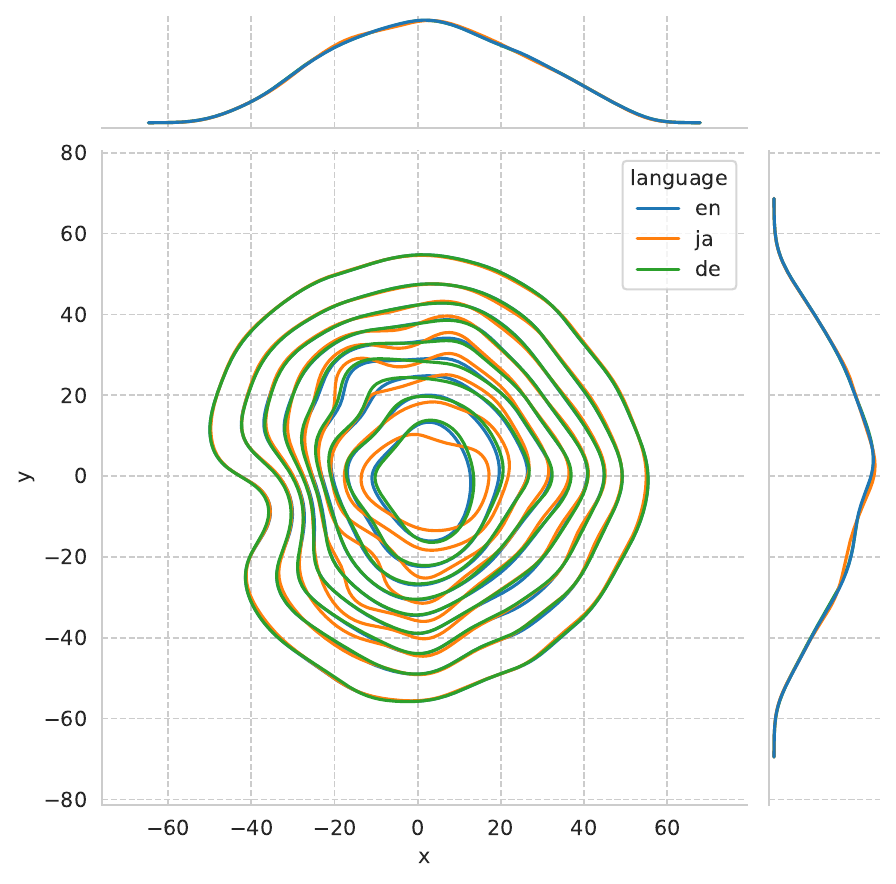}
         \caption{\method}
         \label{fig:visc}
     \end{subfigure}
        \caption{Bivariate kernel density estimation plots of representations after using T-SNE dimensionality reduction to 2 dimension. The blue line is English, the orange line is Japanese and the green line is German. This figure illustrates that the sentence representations are drawn closer after applying \method}
        \label{fig:viss}
\end{figure*}

\subsection{Visualization}
In order to visualize the sentence representations across languages, we retrieve the sentence representation $\mathcal{R}(s)$ for each sentence in \tedm, resulting in 34260 samples in the high-dimensional space. 

To facilitate visualization, we apply T-SNE dimension reduction to reduce the 1024-dim representations to 2-dim. Then we select 3 representative languages: English, German, Japanese and depict the bivariate kernel density estimation based on the 2-dim representations. It is clear in Figure~\ref{fig:viss} that \baselinea cannot align the 3 languages. By contrast, \method draws the representations across 3 languages much closer.

\section{Related Work}
\label{sec:related}
\paragraph{Multilingual Neural Machine Translation}
While initial research on NMT starts with building translation systems between two languages,
\citet{dong2015multi} extends the bilingual NMT to one-to-many translation with sharing encoders across 4 language pairs. 
Hence, there has been a massive increase in work on MT systems that involve more than two languages~\cite{chen2018zero,choi2017improving,chu2018multilingual,dabre2017enabling}.
Recent efforts mainly focuses on designing  language specific components for multilingual NMT to enhance the model performance on rich-resource languages~\cite{bapna2019simple,kim2019effective,wang2019compact,escolano2020training}.
Another promising thread line is to enlarge the model size with extensive training data to improve the model capability~\cite{arivazhagan2019massively,aharoni2019massively,fan2020beyond}.
Different from these approaches, \method proposes to explicitly close the semantic  representation of different languages and make the most of cross lingual transfer.

\paragraph{Zero-shot Machine Translation}
Typical zero-shot machine translation models rely on a pivot language (e.g. English) to combine the source-pivot and pivot-target translation models~\cite{chen2017teacher,ha2017effective,gu2019improved,currey2019zero}.  
~\citet{johnson-etal-2017-googles} shows that a multilingual NMT system enables zero-shot translation without explicitly introducing pivot methods. 
Promising, but the performance still lags behind the pivot competitors.  
Most following up studies focused on data augmentation methods.
~\citet{zhang-2020-improving-massive}  improved the zero-shot translation with online back translation. 
~\citet{ji2020cross,liu2020multilingual} shows that large scale monolingual data can improve the zero-shot translation with unsupervised pre-training.
~\citet{fan2020beyond} proposes a simple and effective data mining method to enlarge the training corpus of zero-shot directions. 
Some work also attempted to explicitly learn shared semantic representation of different languages to improve the zero-shot translation. 
~\citet{neural-interlingua} suggests that by learning an explicit ``interlingual'' across languages, multilingual NMT model can significantly improve zero-shot translation quality. 
~\citet{al2019consistency} introduces a consistent agreement-based training method that encourages the model to produce equivalent translations of parallel sentences in auxiliary languages.
Different from these efforts, \method attempts to learn a universal many-to-many model, and bridge the cross-lingual representation with contrastive learning and m-RAS. The performance is very competitive both on zero-shot and supervised directions on large scale experiments.  

\paragraph{Contrastive Learning}
Contrastive Learning has become a rising domain and achieved significant success in various computer vision tasks~\cite{zhuang2019local,tian2019contrastive,he2020momentum,chen2020simple,misra2020self}.
Researchers in the NLP domain have also explored contrastive Learning for sentence representation. 
~\citet{wu2020clear}  employed multiple
sentence-level augmentation strategies to learn a noise-invariant sentence representation.
~\citet{fang2020cert} applies
the back-translation to create augmentations of
original sentences. 
Inspired by these studies, we apply contrastive learning for multilingual NMT. 

\paragraph{Cross-lingual Representation}
Cross-lingual representation learning has been intensively studied in order to improve cross-lingual understanding (XLU) tasks. Multilingual masked language models (MLM), such as mBERT\cite{delvin2018bert} and XLM\cite{xlm}, train large Transformer models on multiple languages jointly and have built strong benchmarks on XLU tasks. Most of the previous works on cross-lingual representation learning focus on unsupervised training. For supervised learning, \citet{xlm} proposes TLM objective that simply concatenates parallel sentences as input. By contrast, \method leverages the supervision signal by pulling closer the representations of parallel sentences.

\section{Conclusion}
\label{sec:conclusion}
We demonstrate that contrastive learning can significantly improve zero-shot machine translation directions. 
Combined with additional unsupervised monolingual data, we achieve substantial improvements on all translation directions of multilingual NMT. 
We analyze and visualize our method, and find that contrastive learning tends to close the representation gap of different languages. 
Our results also show the possibilities of training a true many-to-many Multilingual NMT that works well on any translation direction. 
In future work, we will scale-up the current training to more languages, e.g. PC150. 
As such, a single model can handle more than 100 languages and outperforms the corresponding bilingual baseline.

\bibliographystyle{acl_natbib}
\bibliography{paper}

\begin{thebibliography}{43}
\expandafter\ifx\csname natexlab\endcsname\relax\def\natexlab#1{#1}\fi

\bibitem[{Aharoni et~al.(2019)Aharoni, Johnson, and
  Firat}]{aharoni2019massively}
Roee Aharoni, Melvin Johnson, and Orhan Firat. 2019.
\newblock \href {https://doi.org/10.18653/v1/n19-1388} {Massively multilingual
  neural machine translation}.
\newblock In \emph{Proceedings of the 2019 Conference of the North American
  Chapter of the Association for Computational Linguistics: Human Language
  Technologies, {NAACL-HLT} 2019, Minneapolis, MN, USA, June 2-7, 2019, Volume
  1 (Long and Short Papers)}, pages 3874--3884. Association for Computational
  Linguistics.

\bibitem[{Al{-}Shedivat and Parikh(2019)}]{al2019consistency}
Maruan Al{-}Shedivat and Ankur~P. Parikh. 2019.
\newblock \href {https://doi.org/10.18653/v1/n19-1121} {Consistency by
  agreement in zero-shot neural machine translation}.
\newblock In \emph{Proceedings of the 2019 Conference of the North American
  Chapter of the Association for Computational Linguistics: Human Language
  Technologies, {NAACL-HLT} 2019, Minneapolis, MN, USA, June 2-7, 2019, Volume
  1 (Long and Short Papers)}, pages 1184--1197. Association for Computational
  Linguistics.

\bibitem[{Arivazhagan et~al.(2019)Arivazhagan, Bapna, Firat, Lepikhin, Johnson,
  Krikun, Chen, Cao, Foster, Cherry, Macherey, Chen, and
  Wu}]{arivazhagan2019massively}
Naveen Arivazhagan, Ankur Bapna, Orhan Firat, Dmitry Lepikhin, Melvin Johnson,
  Maxim Krikun, Mia~Xu Chen, Yuan Cao, George~F. Foster, Colin Cherry, Wolfgang
  Macherey, Zhifeng Chen, and Yonghui Wu. 2019.
\newblock \href {http://arxiv.org/abs/1907.05019} {Massively multilingual
  neural machine translation in the wild: Findings and challenges}.
\newblock \emph{CoRR}, abs/1907.05019.

\bibitem[{Artetxe and Schwenk(2019)}]{tatoeba}
Mikel Artetxe and Holger Schwenk. 2019.
\newblock \href {https://transacl.org/ojs/index.php/tacl/article/view/1742}
  {Massively multilingual sentence embeddings for zero-shot cross-lingual
  transfer and beyond}.
\newblock \emph{Trans. Assoc. Comput. Linguistics}, 7:597--610.

\bibitem[{Bapna and Firat(2019)}]{bapna2019simple}
Ankur Bapna and Orhan Firat. 2019.
\newblock \href {https://doi.org/10.18653/v1/D19-1165} {Simple, scalable
  adaptation for neural machine translation}.
\newblock In \emph{Proceedings of the 2019 Conference on Empirical Methods in
  Natural Language Processing and the 9th International Joint Conference on
  Natural Language Processing, {EMNLP-IJCNLP} 2019, Hong Kong, China, November
  3-7, 2019}, pages 1538--1548. Association for Computational Linguistics.

\bibitem[{Chen et~al.(2020)Chen, Kornblith, Norouzi, and
  Hinton}]{chen2020simple}
Ting Chen, Simon Kornblith, Mohammad Norouzi, and Geoffrey~E. Hinton. 2020.
\newblock \href {http://proceedings.mlr.press/v119/chen20j.html} {A simple
  framework for contrastive learning of visual representations}.
\newblock In \emph{Proceedings of the 37th International Conference on Machine
  Learning, {ICML} 2020, 13-18 July 2020, Virtual Event}, volume 119 of
  \emph{Proceedings of Machine Learning Research}, pages 1597--1607. {PMLR}.

\bibitem[{Chen et~al.(2017)Chen, Liu, Cheng, and Li}]{chen2017teacher}
Yun Chen, Yang Liu, Yong Cheng, and Victor O.~K. Li. 2017.
\newblock \href {https://doi.org/10.18653/v1/P17-1176} {A teacher-student
  framework for zero-resource neural machine translation}.
\newblock In \emph{Proceedings of the 55th Annual Meeting of the Association
  for Computational Linguistics, {ACL} 2017, Vancouver, Canada, July 30 -
  August 4, Volume 1: Long Papers}, pages 1925--1935. Association for
  Computational Linguistics.

\bibitem[{Chen et~al.(2018)Chen, Liu, and Li}]{chen2018zero}
Yun Chen, Yang Liu, and Victor O.~K. Li. 2018.
\newblock \href
  {https://www.aaai.org/ocs/index.php/AAAI/AAAI18/paper/view/16709}
  {Zero-resource neural machine translation with multi-agent communication
  game}.
\newblock In \emph{Proceedings of the Thirty-Second {AAAI} Conference on
  Artificial Intelligence, (AAAI-18), the 30th innovative Applications of
  Artificial Intelligence (IAAI-18), and the 8th {AAAI} Symposium on
  Educational Advances in Artificial Intelligence (EAAI-18), New Orleans,
  Louisiana, USA, February 2-7, 2018}, pages 5086--5093. {AAAI} Press.

\bibitem[{Choi et~al.(2018)Choi, Shin, and Kim}]{choi2017improving}
Gyu{-}Hyeon Choi, Jong{-}Hun Shin, and Young~Kil Kim. 2018.
\newblock \href
  {http://www.lrec-conf.org/proceedings/lrec2018/summaries/139.html} {Improving
  a multi-source neural machine translation model with corpus extension for
  low-resource languages}.
\newblock In \emph{Proceedings of the Eleventh International Conference on
  Language Resources and Evaluation, {LREC} 2018, Miyazaki, Japan, May 7-12,
  2018}. European Language Resources Association {(ELRA)}.

\bibitem[{Chu and Dabre(2019)}]{chu2018multilingual}
Chenhui Chu and Raj Dabre. 2019.
\newblock \href {http://arxiv.org/abs/1906.07978} {Multilingual multi-domain
  adaptation approaches for neural machine translation}.
\newblock \emph{CoRR}, abs/1906.07978.

\bibitem[{Conneau and Lample(2019)}]{xlm}
Alexis Conneau and Guillaume Lample. 2019.
\newblock \href
  {http://papers.nips.cc/paper/8928-cross-lingual-language-model-pretraining}
  {Cross-lingual language model pretraining}.
\newblock In \emph{Advances in Neural Information Processing Systems 32: Annual
  Conference on Neural Information Processing Systems 2019, NeurIPS 2019, 8-14
  December 2019, Vancouver, BC, Canada}, pages 7057--7067.

\bibitem[{Currey and Heafield(2019)}]{currey2019zero}
Anna Currey and Kenneth Heafield. 2019.
\newblock \href {https://doi.org/10.18653/v1/D19-5610} {Zero-resource neural
  machine translation with monolingual pivot data}.
\newblock In \emph{Proceedings of the 3rd Workshop on Neural Generation and
  Translation@EMNLP-IJCNLP 2019, Hong Kong, November 4, 2019}, pages 99--107.
  Association for Computational Linguistics.

\bibitem[{Dabre et~al.(2017)Dabre, Cromier{\`{e}}s, and
  Kurohashi}]{dabre2017enabling}
Raj Dabre, Fabien Cromier{\`{e}}s, and Sadao Kurohashi. 2017.
\newblock \href {http://arxiv.org/abs/1702.06135} {Enabling multi-source neural
  machine translation by concatenating source sentences in multiple languages}.
\newblock \emph{CoRR}, abs/1702.06135.

\bibitem[{Devlin et~al.(2019)Devlin, Chang, Lee, and
  Toutanova}]{delvin2018bert}
Jacob Devlin, Ming{-}Wei Chang, Kenton Lee, and Kristina Toutanova. 2019.
\newblock \href {https://doi.org/10.18653/v1/n19-1423} {{BERT:} pre-training of
  deep bidirectional transformers for language understanding}.
\newblock In \emph{Proceedings of the 2019 Conference of the North American
  Chapter of the Association for Computational Linguistics: Human Language
  Technologies, {NAACL-HLT} 2019, Minneapolis, MN, USA, June 2-7, 2019, Volume
  1 (Long and Short Papers)}, pages 4171--4186. Association for Computational
  Linguistics.

\bibitem[{Dong et~al.(2015)Dong, Wu, He, Yu, and Wang}]{dong2015multi}
Daxiang Dong, Hua Wu, Wei He, Dianhai Yu, and Haifeng Wang. 2015.
\newblock \href {https://doi.org/10.3115/v1/p15-1166} {Multi-task learning for
  multiple language translation}.
\newblock In \emph{Proceedings of the 53rd Annual Meeting of the Association
  for Computational Linguistics and the 7th International Joint Conference on
  Natural Language Processing of the Asian Federation of Natural Language
  Processing, {ACL} 2015, July 26-31, 2015, Beijing, China, Volume 1: Long
  Papers}, pages 1723--1732. The Association for Computer Linguistics.

\bibitem[{Escolano et~al.(2020)Escolano, Costa{-}juss{\`{a}}, Fonollosa, and
  Artetxe}]{escolano2020training}
Carlos Escolano, Marta~R. Costa{-}juss{\`{a}}, Jos{\'{e}} A.~R. Fonollosa, and
  Mikel Artetxe. 2020.
\newblock \href {http://arxiv.org/abs/2006.01594} {Training multilingual
  machine translation by alternately freezing language-specific
  encoders-decoders}.
\newblock \emph{CoRR}, abs/2006.01594.

\bibitem[{Fan et~al.(2020)Fan, Bhosale, Schwenk, Ma, El{-}Kishky, Goyal,
  Baines, Celebi, Wenzek, Chaudhary, Goyal, Birch, Liptchinsky, Edunov, Grave,
  Auli, and Joulin}]{fan2020beyond}
Angela Fan, Shruti Bhosale, Holger Schwenk, Zhiyi Ma, Ahmed El{-}Kishky,
  Siddharth Goyal, Mandeep Baines, Onur Celebi, Guillaume Wenzek, Vishrav
  Chaudhary, Naman Goyal, Tom Birch, Vitaliy Liptchinsky, Sergey Edunov,
  Edouard Grave, Michael Auli, and Armand Joulin. 2020.
\newblock \href {http://arxiv.org/abs/2010.11125} {Beyond english-centric
  multilingual machine translation}.
\newblock \emph{CoRR}, abs/2010.11125.

\bibitem[{Fang and Xie(2020)}]{fang2020cert}
Hongchao Fang and Pengtao Xie. 2020.
\newblock \href {http://arxiv.org/abs/2005.12766} {{CERT:} contrastive
  self-supervised learning for language understanding}.
\newblock \emph{CoRR}, abs/2005.12766.

\bibitem[{Gu et~al.(2019)Gu, Wang, Cho, and Li}]{gu2019improved}
Jiatao Gu, Yong Wang, Kyunghyun Cho, and Victor O.~K. Li. 2019.
\newblock \href {https://doi.org/10.18653/v1/p19-1121} {Improved zero-shot
  neural machine translation via ignoring spurious correlations}.
\newblock In \emph{Proceedings of the 57th Conference of the Association for
  Computational Linguistics, {ACL} 2019, Florence, Italy, July 28- August 2,
  2019, Volume 1: Long Papers}, pages 1258--1268. Association for Computational
  Linguistics.

\bibitem[{Ha et~al.(2017)Ha, Niehues, and Waibel}]{ha2017effective}
Thanh{-}Le Ha, Jan Niehues, and Alexander~H. Waibel. 2017.
\newblock \href {http://arxiv.org/abs/1711.07893} {Effective strategies in
  zero-shot neural machine translation}.
\newblock \emph{CoRR}, abs/1711.07893.

\bibitem[{He et~al.(2020)He, Fan, Wu, Xie, and Girshick}]{he2020momentum}
Kaiming He, Haoqi Fan, Yuxin Wu, Saining Xie, and Ross~B. Girshick. 2020.
\newblock \href {https://doi.org/10.1109/CVPR42600.2020.00975} {Momentum
  contrast for unsupervised visual representation learning}.
\newblock In \emph{2020 {IEEE/CVF} Conference on Computer Vision and Pattern
  Recognition, {CVPR} 2020, Seattle, WA, USA, June 13-19, 2020}, pages
  9726--9735. {IEEE}.

\bibitem[{Ji et~al.(2020)Ji, Zhang, Duan, Zhang, Chen, and Luo}]{ji2020cross}
Baijun Ji, Zhirui Zhang, Xiangyu Duan, Min Zhang, Boxing Chen, and Weihua Luo.
  2020.
\newblock \href {https://aaai.org/ojs/index.php/AAAI/article/view/5341}
  {Cross-lingual pre-training based transfer for zero-shot neural machine
  translation}.
\newblock In \emph{The Thirty-Fourth {AAAI} Conference on Artificial
  Intelligence, {AAAI} 2020, The Thirty-Second Innovative Applications of
  Artificial Intelligence Conference, {IAAI} 2020, The Tenth {AAAI} Symposium
  on Educational Advances in Artificial Intelligence, {EAAI} 2020, New York,
  NY, USA, February 7-12, 2020}, pages 115--122. {AAAI} Press.

\bibitem[{Johnson et~al.(2017)Johnson, Schuster, Le, Krikun, Wu, Chen, Thorat,
  Vi{\'e}gas, Wattenberg, Corrado, Hughes, and
  Dean}]{johnson-etal-2017-googles}
Melvin Johnson, Mike Schuster, Quoc~V. Le, Maxim Krikun, Yonghui Wu, Zhifeng
  Chen, Nikhil Thorat, Fernanda Vi{\'e}gas, Martin Wattenberg, Greg Corrado,
  Macduff Hughes, and Jeffrey Dean. 2017.
\newblock \href {https://doi.org/10.1162/tacl_a_00065} {{G}oogle{'}s
  multilingual neural machine translation system: Enabling zero-shot
  translation}.
\newblock \emph{Transactions of the Association for Computational Linguistics},
  5:339--351.

\bibitem[{Kim et~al.(2019)Kim, Gao, and Ney}]{kim2019effective}
Yunsu Kim, Yingbo Gao, and Hermann Ney. 2019.
\newblock \href {https://doi.org/10.18653/v1/p19-1120} {Effective cross-lingual
  transfer of neural machine translation models without shared vocabularies}.
\newblock In \emph{Proceedings of the 57th Conference of the Association for
  Computational Linguistics, {ACL} 2019, Florence, Italy, July 28- August 2,
  2019, Volume 1: Long Papers}, pages 1246--1257. Association for Computational
  Linguistics.

\bibitem[{Kingma and Ba(2015)}]{adam}
Diederik~P. Kingma and Jimmy Ba. 2015.
\newblock \href {http://arxiv.org/abs/1412.6980} {Adam: {A} method for
  stochastic optimization}.
\newblock In \emph{3rd International Conference on Learning Representations,
  {ICLR} 2015, San Diego, CA, USA, May 7-9, 2015, Conference Track
  Proceedings}.

\bibitem[{Lample et~al.(2018)Lample, Conneau, Denoyer, and
  Ranzato}]{lample2017unsupervised}
Guillaume Lample, Alexis Conneau, Ludovic Denoyer, and Marc'Aurelio Ranzato.
  2018.
\newblock \href {https://openreview.net/forum?id=rkYTTf-AZ} {Unsupervised
  machine translation using monolingual corpora only}.
\newblock In \emph{6th International Conference on Learning Representations,
  {ICLR} 2018, Vancouver, BC, Canada, April 30 - May 3, 2018, Conference Track
  Proceedings}. OpenReview.net.

\bibitem[{Lin et~al.(2020)Lin, Pan, Wang, Qiu, Feng, Zhou, and Li}]{mrasp}
Zehui Lin, Xiao Pan, Mingxuan Wang, Xipeng Qiu, Jiangtao Feng, Hao Zhou, and
  Lei Li. 2020.
\newblock \href {https://doi.org/10.18653/v1/2020.emnlp-main.210} {Pre-training
  multilingual neural machine translation by leveraging alignment information}.
\newblock In \emph{Proceedings of the 2020 Conference on Empirical Methods in
  Natural Language Processing (EMNLP)}, pages 2649--2663, Online. Association
  for Computational Linguistics.

\bibitem[{Liu et~al.(2020)Liu, Gu, Goyal, Li, Edunov, Ghazvininejad, Lewis, and
  Zettlemoyer}]{liu2020multilingual}
Yinhan Liu, Jiatao Gu, Naman Goyal, Xian Li, Sergey Edunov, Marjan
  Ghazvininejad, Mike Lewis, and Luke Zettlemoyer. 2020.
\newblock \href {https://transacl.org/ojs/index.php/tacl/article/view/2107}
  {Multilingual denoising pre-training for neural machine translation}.
\newblock \emph{Trans. Assoc. Comput. Linguistics}, 8:726--742.

\bibitem[{Lu et~al.(2018)Lu, Keung, Ladhak, Bhardwaj, Zhang, and
  Sun}]{neural-interlingua}
Yichao Lu, Phillip Keung, Faisal Ladhak, Vikas Bhardwaj, Shaonan Zhang, and
  Jason Sun. 2018.
\newblock \href {https://doi.org/10.18653/v1/W18-6309} {A neural interlingua
  for multilingual machine translation}.
\newblock In \emph{Proceedings of the Third Conference on Machine Translation:
  Research Papers}, pages 84--92, Brussels, Belgium. Association for
  Computational Linguistics.

\bibitem[{Misra and van~der Maaten(2020)}]{misra2020self}
Ishan Misra and Laurens van~der Maaten. 2020.
\newblock \href {https://doi.org/10.1109/CVPR42600.2020.00674} {Self-supervised
  learning of pretext-invariant representations}.
\newblock In \emph{2020 {IEEE/CVF} Conference on Computer Vision and Pattern
  Recognition, {CVPR} 2020, Seattle, WA, USA, June 13-19, 2020}, pages
  6706--6716. {IEEE}.

\bibitem[{Post(2018)}]{DBLP:conf/wmt/Post18}
Matt Post. 2018.
\newblock \href {https://doi.org/10.18653/v1/w18-6319} {A call for clarity in
  reporting {BLEU} scores}.
\newblock In \emph{Proceedings of the Third Conference on Machine Translation:
  Research Papers, {WMT} 2018, Belgium, Brussels, October 31 - November 1,
  2018}, pages 186--191. Association for Computational Linguistics.

\bibitem[{Sennrich et~al.(2016)Sennrich, Haddow, and Birch}]{bpe}
Rico Sennrich, Barry Haddow, and Alexandra Birch. 2016.
\newblock \href {https://doi.org/10.18653/v1/P16-1162} {Neural machine
  translation of rare words with subword units}.
\newblock In \emph{Proceedings of the 54th ACL (Volume 1: Long Papers)}, pages
  1715--1725, Berlin, Germany. Association for Computational Linguistics.

\bibitem[{Siddhant et~al.(2020)Siddhant, Bapna, Cao, Firat, Chen, Kudugunta,
  Arivazhagan, and Wu}]{siddhant2020leveraging}
Aditya Siddhant, Ankur Bapna, Yuan Cao, Orhan Firat, Mia~Xu Chen, Sneha~Reddy
  Kudugunta, Naveen Arivazhagan, and Yonghui Wu. 2020.
\newblock \href {https://doi.org/10.18653/v1/2020.acl-main.252} {Leveraging
  monolingual data with self-supervision for multilingual neural machine
  translation}.
\newblock In \emph{Proceedings of the 58th Annual Meeting of the Association
  for Computational Linguistics, {ACL} 2020, Online, July 5-10, 2020}, pages
  2827--2835. Association for Computational Linguistics.

\bibitem[{Song et~al.(2019)Song, Tan, Qin, Lu, and Liu}]{song2019mass}
Kaitao Song, Xu~Tan, Tao Qin, Jianfeng Lu, and Tie{-}Yan Liu. 2019.
\newblock \href {http://proceedings.mlr.press/v97/song19d.html} {{MASS:} masked
  sequence to sequence pre-training for language generation}.
\newblock In \emph{Proceedings of the 36th International Conference on Machine
  Learning, {ICML} 2019, 9-15 June 2019, Long Beach, California, {USA}},
  volume~97 of \emph{Proceedings of Machine Learning Research}, pages
  5926--5936. {PMLR}.

\bibitem[{Tan et~al.(2019)Tan, Ren, He, Qin, Zhao, and
  Liu}]{tan2019multilingual}
Xu~Tan, Yi~Ren, Di~He, Tao Qin, Zhou Zhao, and Tie{-}Yan Liu. 2019.
\newblock \href {https://openreview.net/forum?id=S1gUsoR9YX} {Multilingual
  neural machine translation with knowledge distillation}.
\newblock In \emph{7th International Conference on Learning Representations,
  {ICLR} 2019, New Orleans, LA, USA, May 6-9, 2019}. OpenReview.net.

\bibitem[{Tian et~al.(2020)Tian, Krishnan, and Isola}]{tian2019contrastive}
Yonglong Tian, Dilip Krishnan, and Phillip Isola. 2020.
\newblock \href {https://doi.org/10.1007/978-3-030-58621-8\_45} {Contrastive
  multiview coding}.
\newblock In \emph{Computer Vision - {ECCV} 2020 - 16th European Conference,
  Glasgow, UK, August 23-28, 2020, Proceedings, Part {XI}}, volume 12356 of
  \emph{Lecture Notes in Computer Science}, pages 776--794. Springer.

\bibitem[{Tran et~al.(2020)Tran, Tang, Li, and Gu}]{cross-iter}
Chau Tran, Yuqing Tang, Xian Li, and Jiatao Gu. 2020.
\newblock Cross-lingual retrieval for iterative self-supervised training.
\newblock In \emph{Advances in Neural Information Processing Systems 33: Annual
  Conference on Neural Information Processing Systems 2020, NeurIPS 2020,
  December 6-12, 2020, virtual}.

\bibitem[{Vaswani et~al.(2017)Vaswani, Shazeer, Parmar, Uszkoreit, Jones,
  Gomez, Kaiser, and Polosukhin}]{vaswani2017}
Ashish Vaswani, Noam Shazeer, Niki Parmar, Jakob Uszkoreit, Llion Jones,
  Aidan~N Gomez, {\L}ukasz Kaiser, and Illia Polosukhin. 2017.
\newblock {Attention is All you Need}.
\newblock In \emph{Advances in Neural Information Processing Systems},
  volume~30, pages 5998--6008. Curran Associates, Inc.

\bibitem[{Wang et~al.(2019{\natexlab{a}})Wang, Li, Xiao, Zhu, Li, Wong, and
  Chao}]{deep-transformer}
Qiang Wang, Bei Li, Tong Xiao, Jingbo Zhu, Changliang Li, Derek~F. Wong, and
  Lidia~S. Chao. 2019{\natexlab{a}}.
\newblock \href {https://doi.org/10.18653/v1/P19-1176} {Learning deep
  transformer models for machine translation}.
\newblock In \emph{Proceedings of the 57th ACL}, pages 1810--1822, Florence,
  Italy. Association for Computational Linguistics.

\bibitem[{Wang et~al.(2019{\natexlab{b}})Wang, Zhou, Zhang, Zhai, Xu, and
  Zong}]{wang2019compact}
Yining Wang, Long Zhou, Jiajun Zhang, Feifei Zhai, Jingfang Xu, and Chengqing
  Zong. 2019{\natexlab{b}}.
\newblock \href {https://doi.org/10.18653/v1/p19-1117} {A compact and
  language-sensitive multilingual translation method}.
\newblock In \emph{Proceedings of the 57th Conference of the Association for
  Computational Linguistics, {ACL} 2019, Florence, Italy, July 28- August 2,
  2019, Volume 1: Long Papers}, pages 1213--1223. Association for Computational
  Linguistics.

\bibitem[{Wu et~al.(2020)Wu, Wang, Gu, Khabsa, Sun, and Ma}]{wu2020clear}
Zhuofeng Wu, Sinong Wang, Jiatao Gu, Madian Khabsa, Fei Sun, and Hao Ma. 2020.
\newblock \href {http://arxiv.org/abs/2012.15466} {{CLEAR:} contrastive
  learning for sentence representation}.
\newblock \emph{CoRR}, abs/2012.15466.

\bibitem[{Zhang et~al.(2020)Zhang, Williams, Titov, and
  Sennrich}]{zhang-2020-improving-massive}
Biao Zhang, Philip Williams, Ivan Titov, and Rico Sennrich. 2020.
\newblock \href {https://doi.org/10.18653/v1/2020.acl-main.148} {Improving
  massively multilingual neural machine translation and zero-shot translation}.
\newblock In \emph{Proceedings of the 58th ACL}, pages 1628--1639, Online.
  Association for Computational Linguistics.

\bibitem[{Zhuang et~al.(2019)Zhuang, Zhai, and Yamins}]{zhuang2019local}
Chengxu Zhuang, Alex~Lin Zhai, and Daniel Yamins. 2019.
\newblock \href {https://doi.org/10.1109/ICCV.2019.00610} {Local aggregation
  for unsupervised learning of visual embeddings}.
\newblock In \emph{2019 {IEEE/CVF} International Conference on Computer Vision,
  {ICCV} 2019, Seoul, Korea (South), October 27 - November 2, 2019}, pages
  6001--6011. {IEEE}.

\end{thebibliography}

\newpage

\appendix
\label{sec:appendix}
\section{Case Study}

\begin{figure*}[t]
     \begin{center}
     \includegraphics[width=0.6\textwidth]{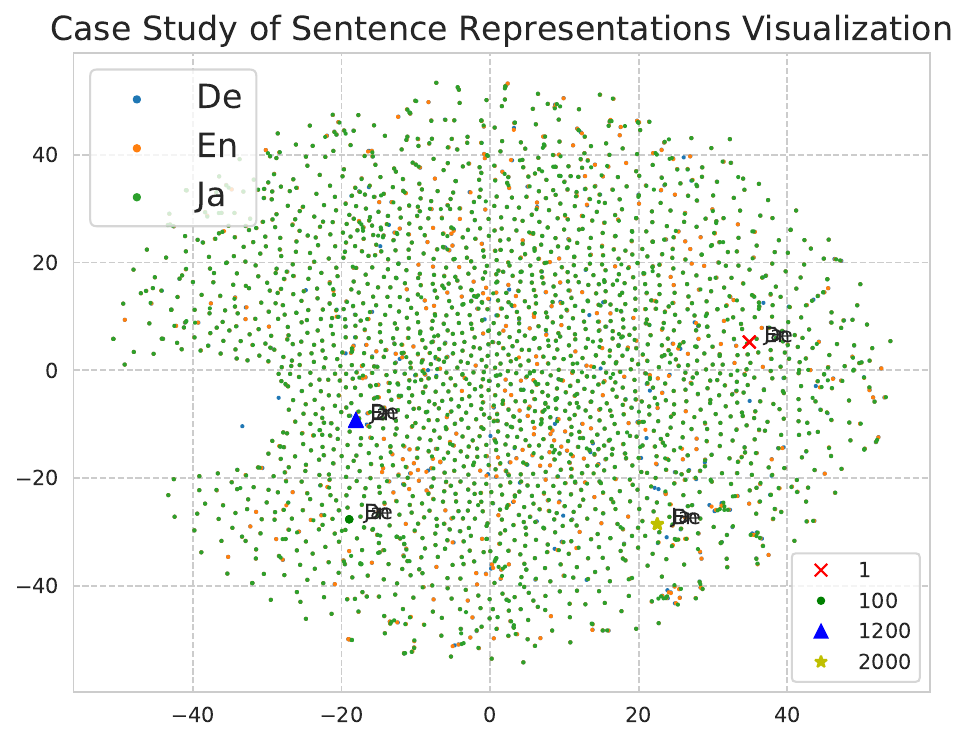}
     \caption{Case Study: Examples of representations of multi-way parallel sentences on \method representation space. We can observe that similar sentences overlap perfectly on the space. Numbers in the legend means the id of sentence in \tedm (See Table~\ref{tab:case} for detailed sentences). We can clearly observe that similar sentences are clustered to the neighboring location.}
     \label{fig:tedplot}
     \end{center}
\end{figure*}

\begin{table*}[hp]
\begin{center}
\resizebox{0.8\linewidth}{!}{
\begin{tabular}{c|c|l}
\toprule
Id & Language & Sentence \\ \hline
& De & Was sie alle eint, ist, dass sie sterben werden.                              \\
1  & En &  The one thing that all of them have in common is that they're going to die.     \\
 & Ja &  \begin{CJK}{UTF8}{min}
１ つ 全員 に 共通 し て 言え る の は 皆 いずれ 死 ぬ と い う こと で す
\end{CJK}  \\
\midrule
& De &  Rechts seht Ihr meinen Kollegen Sören , der sich wirklich in dem Raum befindet.  \\
100 & En & On the right side you can see my colleague Soren , who 's actually in the space. \\
  & Ja & 
\begin{CJK}{UTF8}{min}
右側 に は 同僚 ・ ソーレン が 見え ま す 実際 その 場所 に い た の で す 
\end{CJK} \\
\bottomrule
\end{tabular}
}
\caption{Case Study: Parallel sentences distributed in English, German and Japanese. }
\label{tab:case}
\end{center}
\end{table*}

We plot the location of multi-way parallel sentences in the representation space of \method in Figure~\ref{fig:tedplot} and list sentences number 1 and 100 in Table~\ref{tab:case}


\section{Details of Evaluation Results}
We list detailed results of evaluation on a wide range of test sets.

\subsection{Results on OPUS-100}
\begin{table*}[htp]
\begin{center}
\resizebox{\linewidth}{!}{
\begin{tabular}{c|llllll|r||c|llllll|r}
\toprule
\multicolumn{8}{c}{\textbf{\baselinea}}                     &  \multicolumn{8}{c}{\textbf{\methodb}}                               \\ \hline
    & Ar  & Zh   & Nl  & Fr   & De   & Ru   & \multicolumn{1}{c}{Avg}   &     & Ar  & Zh   & Nl  & Fr   & De   & Ru   & \multicolumn{1}{c}{Avg}  \\ \hline
Ar  & -   & 9.2  & 1.2 & 7.6  & 1.8  & 8.2  & 5.6  &  Ar  & -   & 26.1 & 1.2 & 19.1 & 10.5 & 12.5 & 13.9 \\
Zh  & 4.7 & -    & 0.8 & 7.7  & 1.7  & 5.8  & 4.1  &  Zh  & 5.6 & -    & 0.9 & 32.1 & 8.0  & 17.3 & 12.8 \\
Nl  & 1.9 & 5.1  & -   & 10.8 & 9.9  & 3.7  & 6.3  &  Nl  & 2.3 & 5.5  & -   & 10.3 & 10.3 & 3.8  & 5.6  \\
Fr  & 3.9 & 6.5  & 3.7 & -    & 4.3  & 5.3  & 4.8  &  Fr  & 5.6 & 41.5 & 3.7 & -    & 18.0 & 19.5 & 18.8 \\
De  & 3.1 & 4.4  & 4.5 & 6.5  & -    & 5.5  & 4.8  &  De  & 4.6 & 19.9 & 4.4 & 23.0 & -    & 13.6 & 13.1 \\
Ru  & 4.8 & 8.4  & 1.5 & 5.9  & 3.2  & -    & 4.8  &  Ru  & 5.9 & 37.4 & 1.5 & 30.1 & 12.2 & -    & 17.4 \\
\hline
Avg & 3.7 & 6.7  & 2.3 & 7.7  & 4.2  & 5.7  & \textbf{5.05}    &  Avg & 4.8 & 26.1 & 2.3 & 22.9 & 11.8 & 13.3 &  \textbf{13.55}  \\
    
    \midrule
\multicolumn{8}{c}{\textbf{\methodl}}                       &  \multicolumn{8}{c}{\textbf{\methodx}}                       \\ \hline
    & Ar  & Zh   & Nl  & Fr   & De   & Ru   & \multicolumn{1}{c}{Avg}  &      & Ar  & Zh   & Nl  & Fr   & De   & Ru   & \multicolumn{1}{c}{Avg}  \\ 
    \hline
Ar      & -   & 5.7  & 1.6  & 1.2  & 6.8  & 4.0    & 3.9     &  Ar      & -   & 28.8 & 1.0   & 20.9 & 7.9  & 15.6 & 14.8    \\
Zh      & 4.0   & -    & 4.2  & 3.8  & 5.4  & 2.9  & 4.1     &  Zh      & 6.3 & -    & 0.7 & 33.8 & 5.9  & 20.0   & 13.3    \\
Nl      & 3.0   & 7.5  & -    & 4.4  & 7.8  & 2.6  & 5.1     &  Nl      & 3.2 & 8.1  & -   & 16.3 & 14.3 & 6.0    & 9.6     \\
Fr      & 2.8 & 14.8 & 13.3 & -    & 5.4  & 7.4  & 6.4     &  Fr      & 6.6 & 41.5 & 3.7 & -    & 16.7 & 21.4 & 19.1    \\
De      & 5.3 & 6.1  & 2.5  & 1.4  & -    & 3.4  & 3.7     &  De      & 6.1 & 21.3 & 4.6 & 24.3 & -    & 15.0   & 14.3    \\
Ru      & 5.2 & 6.7  & 1.5  & 1.0    & 5.6  & -    & 4.0     &  Ru      & 7.1 & 38.0   & 1.1 & 30.6 & 11.1 & -    & 17.6    \\
\hline

Avg & 4.1 & 8.2  & 4.6 & 2.4  & 6.2  & 4.1  & \textbf{4.91}  &  Avg & 5.9 & 27.5 & 2.2 & 25.2 & 11.2 & 15.6 & \textbf{14.60} \\
    
    \midrule
\multicolumn{8}{c}{\textbf{\method}}                       &  \multicolumn{8}{c}{\textbf{Pivot}}                       \\ \hline
& Ar  & Zh   & Nl  & Fr   & De   & Ru   & \multicolumn{1}{c}{Avg}  &      & Ar  & Zh   & Nl  & Fr   & De   & Ru   & \multicolumn{1}{c}{Avg}  \\ 
\hline
Ar  & -    & 32.5  & 3.2  & 22.8  & 11.2  & 16.7  & 17.3    &  Ar & -   & 31.4 & 1.0   & 22.9 & 13.5 & 16.4 & 17.0    \\
Zh &  6.5  & -     & 1.9  & 32.9  & 7.6   & 23.7  & 14.5    &  Zh & 7.3 & -    & 0.8 & 37.7 & 11.9 & 24.2 & 16.4    \\
Nl & 1.7  & 8.2   & -    & 7.5   & 10.2  & 2.9   & 6.1    &  Nl & 1.7 & 4.9  & -   & 10.1 & 9.7  & 3.7  & 6.0     \\
Fr & 6.2  & 42.3  & 7.5  & -     & 18.9  & 24.4  & 21.7    &  Fr & 6.8 & 44.1 & 3.6 & -    & 21.4 & 23.2 & 22.3    \\
De & 4.9  & 21.6  & 9.2  & 24.7  & -     & 14.4  & 15.0    &  De & 4.9 & 20.8 & 4.3 & 25.3 & -    & 15.5 & 14.2    \\
Ru & 7.1  & 40.6  & 4.5  & 29.9  & 13.5  & -     & 19.1  &  Ru & 6.7 & 41.5 & 1.4 & 34.5 & 15.5 & -    & 19.9    \\
\hline
Avg & 5.3  & 29.0  & 5.3  & 23.6  & 12.3  & 16.4 & \textbf{15.31} &  Avg & 5.5 & 28.5 & 2.2 & 26.1 & 14.4 & 16.6 & \textbf{15.56}
    
\\ \bottomrule

\end{tabular}
}
\end{center}
\caption{Detailed de-tokenized BLEU on OPUS-100 zero-shot test set. Note that results of \methodl are computed without fine-tuning. }
\label{tab:opusall}
\end{table*}
Detailed results on OPUS-100 zero-shot evaluation set are listed in Table~\ref{tab:opusall}

\subsection{Results on WMT}
\begin{table*}[ht]
\begin{center}

\resizebox{\linewidth}{!}{
\begin{tabular}{lllllllllll|rr}
\toprule
 &
  \multicolumn{2}{c}{\begin{tabular}[c]{@{}c@{}}En-Fr\\ wmt14\end{tabular}} &
  \multicolumn{2}{c}{\begin{tabular}[c]{@{}c@{}}En-De\\ wmt14\end{tabular}} &
  \multicolumn{2}{c}{\begin{tabular}[c]{@{}c@{}}En-Zh\\ wmt17\end{tabular}} &
  \multicolumn{2}{c}{\begin{tabular}[c]{@{}c@{}}En-Ro\\ wmt16\end{tabular}} &
  \multicolumn{2}{c}{\begin{tabular}[c]{@{}c@{}}En-Cs\\ wmt16\end{tabular}} &
   &
   \\
 &
  \multicolumn{1}{c}{$\rightarrow$} &
  \multicolumn{1}{c}{$\leftarrow$} &
  \multicolumn{1}{c}{$\rightarrow$} &
  \multicolumn{1}{c}{$\leftarrow$} &
  \multicolumn{1}{c}{$\rightarrow$} &
  \multicolumn{1}{c}{$\leftarrow$} &
  \multicolumn{1}{c}{$\rightarrow$} &
  \multicolumn{1}{c}{$\leftarrow$} &
  \multicolumn{1}{c}{$\rightarrow$} &
  \multicolumn{1}{c}{$\leftarrow$} &
   &
   \\ \hline
\baselinea &
  42.0 & 38.1 & 27.1 & 34.2 & 32.8 & 24.2 & 26.9 & 37.7 & 20.9 & 31.3  &
   &
   \\
\methodb &
  42.1          & 38.7          & 26.8          & 34.6          & 33.2          & 24.7          & 26.6          & 37.5          & 20.8          & 31.5  &
   &
   \\
   \methodl  & 43.1 & 39.2 & 29.2 & 34.6 & 34.8 & 24.8 & 28.2 & 38.8 & 22.5 & 32.1 &       &      \\
\methodx &
  43.3          & 39.3          & 29.1          & 34.7          & \textbf{35.0}          & \textbf{24.5}          & 28.4          & 39.0          & 22.4          & 32.5 &
   &
   \\
\method &
  \textbf{43.5} & \textbf{39.3} & \textbf{29.7} & \textbf{35.0} & 34.6          & 23.8          & \textbf{28.7} & \textbf{39.1} & \textbf{24.3} & \textbf{33.1} &
   &
   \\
   \midrule
 &
  \multicolumn{2}{c}{\begin{tabular}[c]{@{}c@{}}En-Tr\\ wmt16\end{tabular}} &
  \multicolumn{2}{c}{\begin{tabular}[c]{@{}c@{}}En-Ru\\ wmt19\end{tabular}} &
  \multicolumn{2}{c}{\begin{tabular}[c]{@{}c@{}}En-Fi\\ wmt17\end{tabular}} &
  \multicolumn{2}{c}{\begin{tabular}[c]{@{}c@{}}En-Es\\ wmt13\end{tabular}} &
  \multicolumn{2}{c}{\begin{tabular}[c]{@{}c@{}}En-It\\ wmt09\end{tabular}} &
  \multicolumn{1}{c}{Avg} &
  \multicolumn{1}{c}{$\Delta$} \\
 &
  \multicolumn{1}{c}{$\rightarrow$} &
  \multicolumn{1}{c}{$\leftarrow$} &
  \multicolumn{1}{c}{$\rightarrow$} &
  \multicolumn{1}{c}{$\leftarrow$} &
  \multicolumn{1}{c}{$\rightarrow$} &
  \multicolumn{1}{c}{$\leftarrow$} &
  \multicolumn{1}{c}{$\rightarrow$} &
  \multicolumn{1}{c}{$\leftarrow$} &
  \multicolumn{1}{c}{$\rightarrow$} &
  \multicolumn{1}{c}{$\leftarrow$} &
   &
   \\ \hline
\baselinea &
  18.2          & 24.3          & 17.0          & 22.6          & 20.0          & 28.2          & 32.8          & 33.7          & 29.0          & 32.0          & 28.65 &
   \\
\methodb &
  18.2          & 24.8          & 17.6          & 23.2          & 20.0          & 27.8          & 33.1          & 33.2          & 29.2          & 32.2          & 28.79 &
  +0.14 \\
  \methodl & 20.0 & 25.2 & 18.6 & 23.3 & 22.0 & 29.2 & 34.0 & 34.3 & 30.1 & 32.4 & 29.82 & +1.17 \\
\methodx &
  20.4          & 25.7          & 18.6          & \textbf{23.4} & 22.0          & 29.4          & 34.1          & 34.3          & 30.4          & 32.6          & 29.96 &
  +1.31 \\
\method &
  \textbf{21.4} & \textbf{25.8} & \textbf{19.2} & 23.2          & \textbf{23.4} & \textbf{30.1} & \textbf{34.5} & \textbf{35.0} & \textbf{30.8} & \textbf{32.6} & \textbf{30.36} &
  \textbf{+1.71} \\
  \bottomrule
\end{tabular}
}
\caption{Tokenized BLEU score on public WMT testsets. \methodb only adopt contrastive learning on the basis of \baselinea. \methodl excludes \mdataset and contrastive loss from \method. \methodx excludes monolingual data from \method. Note that results of \methodl are computed without fine-tuning.  }
\label{tab:allscore}
\end{center}
\end{table*}
Detailed results on WMT evaluation set are listed in Table~\ref{tab:allscore}


\section{Example of AA}

\begin{figure*}[t]
     \begin{center}
     \includegraphics[width=\textwidth]{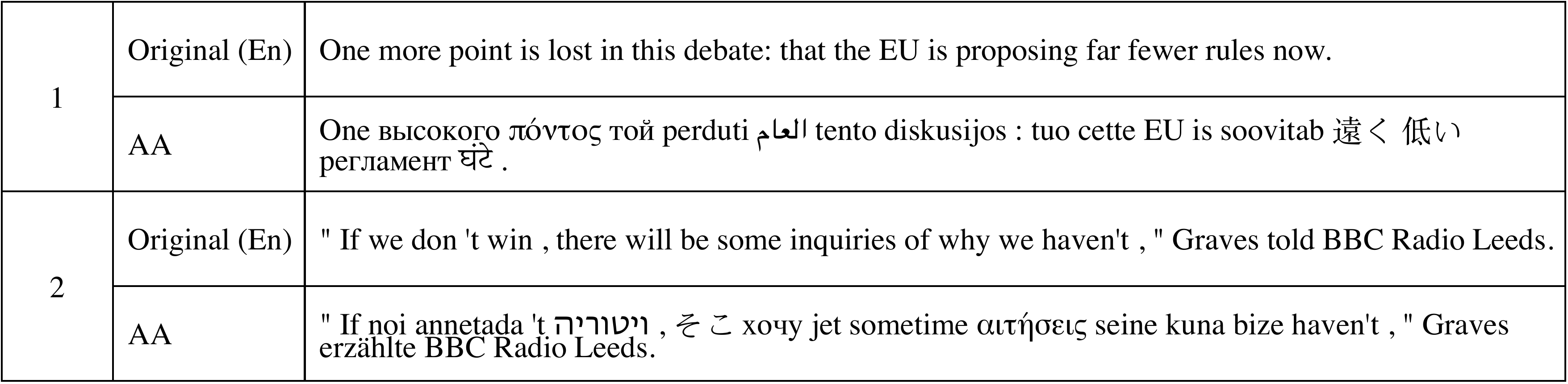}
     \caption{Two examples of sentences with its noised version after AA}
     \label{fig:mras}
     \end{center}
\end{figure*}

We show two results of sentences after AA in Figure~\ref{fig:mras}

\section{Details of \mdataset}
\begin{table*}[htp]
\begin{center}
  
\begin{tabular}{lllll}
\toprule
Lanuage & Original Num. & Sampling Ratio & \% of replaced tokens & Final Num. \\
\hline
bg      & 37870628      & 1.58           & /                     & 59839631   \\
cs      & 75808960      & 0.89           & 0.29                  & 67118121   \\
de      & 319938740     & 0.29           & 0.40                  & 91985353   \\
el      & 4178943       & 5.50           & 0.35                  & 22980970   \\
en      & 224446700     & 0.38           & 0.62                  & 85785847   \\
es      & 17632409      & 1.24           & 0.60                  & 21783966   \\
et      & 4978345       & 7.82           & 0.28                  & 38925275   \\
fi      & 19954908      & 2.57           & 0.29                  & 51368970   \\
fr      & 85274195      & 0.84           & 0.54                  & 71760116   \\
gu      & 530747        & 35.26          & /                     & 18716499   \\
hi      & 6240797       & 1.85           & 0.46                  & 11521321   \\
it      & 39170950      & 1.56           & 0.47                  & 61064797   \\
ja      & 3250665       & 11.14          & 0.15                  & 36225302   \\
kk      & 1853728       & 18.30          & /                     & 33926819   \\
lt      & 2446627       & 13.02          & 0.16                  & 31857781   \\
lv      & 10942229      & 4.30           & 0.35                  & 47032289   \\
ro      & 20094801      & 2.62           & 0.34                  & 52685562   \\
ru      & 89373208      & 0.79           & 0.29                  & 70839964   \\
sr      & 3801560       & 10.30          & /                     & 39167541   \\
tr      & 16337598      & 3.03           & 0.29                  & 49502982   \\
zh      & 4238918       & 8.66           & 0.15                  & 36706289   \\
nl      & 1177713       & 1.00           & 0.52                  & 1177713    \\
pl      & 3404714       & 1.00           & ?                     & 3404714    \\
pt      & 9103090       & 1.00           & ?                     & 9103090    \\
\hline
SUM     &               &                &                       & 1014480912 \\

\bottomrule

\end{tabular}
\caption{Detail of \mdataset, '?' means the data is missing, and '/' means the corresponding language is not contained in the synonym dictionary.}
\label{tab:mdata}
\end{center}
\end{table*}

We describe the detail of \mdataset in Table~\ref{tab:mdata}

\end{document}